\documentclass{article}

\usepackage[preprint]{neurips_2021}

\usepackage[utf8]{inputenc} % allow utf-8 input
\usepackage[T1]{fontenc}    % use 8-bit T1 fonts
\usepackage{hyperref}       % hyperlinks
\usepackage{url}            % simple URL typesetting
\usepackage{booktabs}       % professional-quality tables
\usepackage{amsfonts}       % blackboard math symbols
\usepackage{nicefrac}       % compact symbols for 1/2, etc.
\usepackage{microtype}      % microtypography
\usepackage{xcolor}         % colors
\usepackage{subfigure}
\usepackage{multirow}
\usepackage{graphicx}
\usepackage{ulem}
\usepackage{amsmath}
\usepackage{amssymb,amsthm}

\newtheorem{theorem}{Theorem}

\usepackage{listings}

\usepackage{xcolor}

\definecolor{codegreen}{rgb}{0,0.6,0}
\definecolor{codegray}{rgb}{0.5,0.5,0.5}
\definecolor{codepurple}{rgb}{0.58,0,0.82}
\definecolor{backcolour}{rgb}{0.95,0.95,0.92}

\lstdefinestyle{mystyle}{
  backgroundcolor=\color{backcolour},   commentstyle=\color{codegreen},
  keywordstyle=\color{magenta},
  numberstyle=\tiny\color{codegray},
  stringstyle=\color{codepurple},
  basicstyle=\ttfamily\footnotesize,
  breakatwhitespace=false,         
  breaklines=true,                 
  captionpos=b,                    
  keepspaces=true,                 
  numbers=left,                    
  numbersep=5pt,                  
  showspaces=false,                
  showstringspaces=false,
  showtabs=false,                  
  tabsize=2
}
\usepackage{tikz}
\usepackage{comment}
\usepackage{amsmath,amssymb} % define this before the line numbering.
\usepackage{color}

\usepackage{fancyvrb}
\usepackage{hyperref}
\usepackage[T1]{fontenc}    % use 8-bit T1 fonts
\usepackage{hyperref}       % hyperlinks
\usepackage{nicefrac}       % compact symbols for 1/2, etc.
\usepackage{microtype}      % microtypography
\usepackage{xcolor}         % colors
\usepackage{subfigure}
\usepackage{multirow}
\usepackage{graphicx}

\usepackage{array}
\usepackage{floatrow}
\newfloatcommand{capbtabbox}{table}[][\FBwidth]

\DeclareMathOperator{\Avg}{Avg}
\DeclareMathOperator{\Rad}{Rad}

\newcommand{\cC}{\mathcal{C}}
\newcommand{\cD}{\mathcal{D}}
\newcommand{\cE}{\mathcal{E}}
\newcommand{\cF}{\mathcal{F}}
\newcommand{\cG}{\mathcal{G}}
\newcommand{\cH}{\mathcal{H}}

\newcommand{\cU}{\mathcal{U}}
\newcommand{\cV}{\mathcal{V}}
\newcommand{\cX}{\mathcal{X}}
\newcommand{\cY}{\mathcal{Y}}

\newcommand{\E}{\mathbb{E}}
\newcommand{\R}{\mathbb{R}}
\renewcommand{\P}{\mathbb{P}}

\newcommand{\bI}{\mathbb{I}}
\newcommand{\Var}{\mathrm{Var}}
\DeclareMathOperator*{\argmax}{arg\,max}
\DeclareMathOperator*{\argmin}{arg\,min}

\newcommand{\tx}{\tilde{x}}

\newcommand{\tS}{\tilde{S}}

\newcommand{\tcE}{\tilde{\cE}}

\newcommand{\tP}{\tilde{P}}

\newcommand{\tig}{\tilde{g}}
\newcommand{\tie}{\tilde{e}}
\newcommand{\tiy}{\tilde{y}}
\newcommand{\tih}{\tilde{h}}

\newcommand{\hP}{\hat{P}}
\newcommand{\hiy}{\hat{y}}

\newcommand*\samethanks[1][\value{footnote}]{\footnotemark[#1]}

\usepackage[accsupp]{axessibility}  % Improves PDF readability for those with disabilities.

\definecolor{ForestGreen}{RGB}{34,139,34}

\title{Image2Point: 3D Point-Cloud Understanding with 2D Image Pretrained Models}

\author{
Chenfeng Xu  $^1$\thanks{Equal contribution}
\And Shijia Yang  $^1$\samethanks \And Tomer Galanti  $^2$  \And Bichen Wu  $^3$\thanks{Corresponding author} \And Xiangyu Yue  $^1$\And Bohan Zhai  $^1$ \And Wei Zhan $^1$\And Peter Vajda $^3$ \And Kurt Keutzer $^1$ \And Masayoshi Tomizuka $^1$

\\[0.2cm]
$^1$ University of California, Berkeley, $^2$ Massachusetts Institute of Technology, $^3$ Meta Reality Labs\\

\\ [0.1cm]
\{xuchenfeng, shijiayang, xyyue, zhaibohan, wzhan, keutzer, tomizuka\}@berkeley.edu, \\
galanti@mit.edu, \{wbc, vajdap\}@fb.com\\

  \footnotetext[1]{Equal contribution}
}

\begin{document}

\maketitle

\begin{abstract}
3D point-clouds and 2D images are different visual representations of the physical world. While human vision can understand both representations, computer vision models designed for 2D image and 3D point-cloud understanding are quite different.
Our paper explores the potential of transferring 2D model architectures and weights to understand 3D point-clouds, by empirically investigating the feasibility of the transfer, the benefits of the transfer, and shedding light on why the transfer works.
We discover that we can indeed use the same architecture and pretrained weights of a neural net model to understand both images and point-clouds.
Specifically, we transfer the image-pretrained model to a point-cloud model by copying or inflating the weights.
We find that \textbf{f}inetuning the transformed \textbf{i}mage-\textbf{p}retrained models (FIP) with minimal efforts --- only on input, output, and normalization layers --- can achieve competitive performance on 3D point-cloud classification, beating a wide range of point-cloud models that adopt task-specific architectures and use a variety of tricks. When finetuning the whole model, the performance improves even further. Meanwhile, FIP improves data efficiency, reaching up to 10.0 top-1 accuracy percent on few-shot classification. It also speeds up the training of point-cloud models by up to 11.1x for a target accuracy (e.g., 90 \% accuracy). Lastly, we provide an explanation of the image to point-cloud transfer from the aspect of \textit{neural collapse}. The code is available at:  \url{https://github.com/chenfengxu714/image2point}.
\end{abstract}

\section{Introduction}

Point-cloud is an important visual representation for 3D computer vision. It is widely used in a variety of applications, including autonomous driving~\cite{behley2019iccv,caesar2020nuscenes,yue2018lidar}, robotics~\cite{armeni2017joint,pomerleau2015review,xu2021you}, augmented and virtual reality~\cite{3dwarehouse,Wu_2015_CVPR,7273863}, \textit{etc.} However, a point-cloud represents visual information in a significantly different way from a 2D image. Specifically, a point-cloud consists of a set of unordered points lying on the object's surface, with each point encoding its spatial $x,y,z$ coordinates and potentially other features such as intensity. In contrast, a 2D image organizes visual features as a dense 2D RGB pixel array.
Due to the representation differences, 2D image and 3D point-cloud understanding are treated as two separate problems. 2D image models and point-cloud models are designed to have different architectures and are trained on different types of data. No efforts have tried to directly transfer models from images to point-clouds.

% \begin{figure*}[!t]
%     \centering
%     \includegraphics[width=0.95\textwidth]{figures/image_preforpc_intro1-cropped.pdf}
%     \caption{{\bf Examples of images and point-clouds.} Images are dense, RGB, pixel arrays. Point-clouds are sparse, XYZ, point sets. }
%     \label{fig:intro_pc}
% \end{figure*}

Intuitively, both 3D point-clouds and 2D images are visual representations of the physical world. Their low-level representations are drastically different, but they can represent the same underlying visual concept. Furthermore, human vision has no problem understanding both representations. To connect images and point-clouds, previous works attempted to generate pseudo point-clouds by estimating the depth of mono/stereo images~\cite{wang2019pseudo,gur2019single,yin2021virtual}. However, depth estimation from a single image is a challenging problem in computer vision, which requires large-scale dense depth labels~\cite{Ranftl2020}. Estimating depth from stereo images is easier but requires strict calibrated and synchronized stereo cameras, which limits the data scale. Therefore, it is interesting to ask whether we could use large-scale image models that were pretrained using supervised classification datasets (\textit{e.g., ImageNet1K/ImageNet21K} classification) for point-cloud understanding.

\begin{figure*}[!t]
    \centering
    \includegraphics[width=1\textwidth]{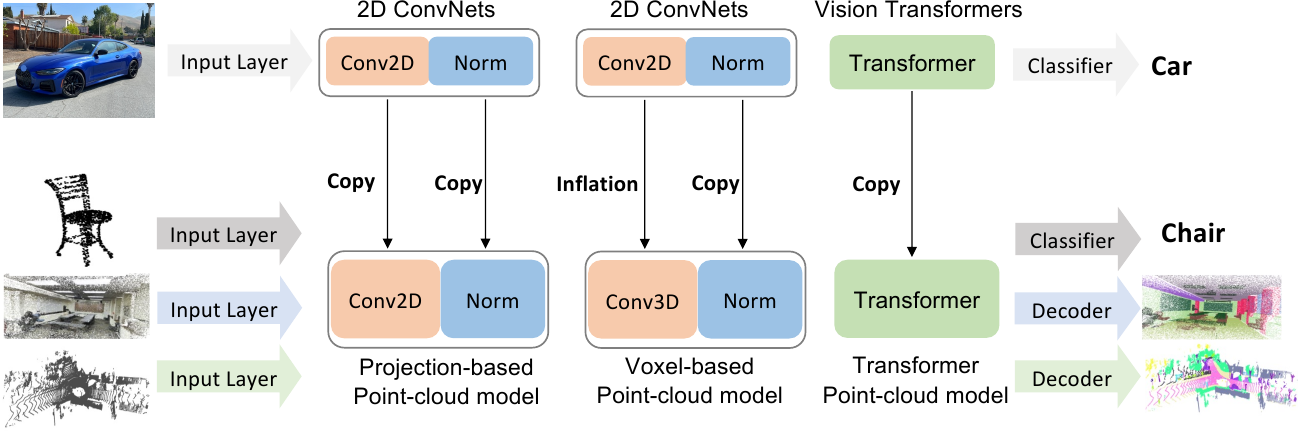}
    \caption{We investigate the feasibility of converting pretrained 2D image models to 3D point-cloud models. For example, with filter inflation and finetuning the input, output (classifier for classification task and decoder for semantic segmentation task), and normalization layers, the transformed 2D ConvNets are able to deal with point-cloud classification, indoor, and driving scene segmentation.}
    \label{fig:intro}
\end{figure*}

Remarkably, the answer to the question above is positive. As we show in this work, 2D image models trained on image datasets can be transferred to understand 3D point-clouds with minimal effort. As illustrated in Figure~\ref{fig:intro}, given the commonly-used image-pretrained models, such as 2D ConvNets \cite{he2016deep} and vision transformers \cite{dosovitskiy2020image}, we can easily convert them into various kinds of point-cloud models. In particular, a pretrained 2D ConvNet and vision transformer can be easily extended into projection-based, voxel-based, and transformer-based point-cloud models via copying weights or inflating weights~\cite{carreira2017quo}.

In this paper, we primarily focus on 3D ConvNets inflated from 2D pre-trained models. With the transformed point-cloud model (\textit{e.g.}, inflated 3D ConvNets), we add linear input and output layers to the network; and on a target point-cloud dataset, we only finetune the input and output layers, and batch normalization layers, while keeping the pretrained model weights untouched. We call such partially-finetuned-image-pretrained models as \textit{FIP-IO+BN} (finetuning input, output, and BN layers). As we show, FIP-IO+BN can achieve competitive performance up to 90.8\% top-1 accuracy on the ModelNet 3D Warehouse dataset, on top of ResNet50, outperforming previous point-cloud models that adopt task-specific model architectures and tricks. 

Even though incorporating pretrained models is useful for tackling downstream tasks, point-cloud models are typically trained from scratch. Based on our discovery, we further investigate fully-finetuned-image-pretrained models (termed as \textit{FIP-ALL}). We observe that FIP-ALL brings significant improvement on top of different kinds of point-cloud models transformed from image-pretrained models. Besides, we also find that it generalizes to PointNet++ \cite{qi2017pointnetplusplus} which is pre-trained on images by ourselves. Specifically, FIP-ALL outperforms the training-from-scratch by a large margin on top of PointNet++, SimpleView, ViT-B-16, and ViT-L-16, respectively. In addition to the performance gain, FIP-ALL exhibits superior data efficiency with up to $10.0\%$ accuracy improvement in few-shot classification on the ModelNet 3D Warehouse dataset. Compared with training-from-scratch, FIP-ALL also dramatically speeds up the training by using 11.1 times fewer epochs to reach the same validation accuracy (\textit{e.g.}, 90\% accuracy).

% Finally, we provide theoretical evidence for the success of transfer learning between two classification tasks of different modalities. The analysis is based on an extension of the framework proposed in~\cite{galanti2022on}. 

Finally, we theoretically explore the relationship between transferring knowledge between tasks of different modalities and neural collapse to shed light on why the transfer works. The analysis is based on extending the framework proposed in~\cite{galanti2022on} and is provided in Appendix \ref{spp:theory}. 

%Neural collapse show a phenomenon that the features learned by overparameterized classification networks show an interesting clustering property. We demonstrate both theoretically and empirically that neural collapse generalizes to image-to-point too. 

\section{Related Work}

\subsection{Point-Cloud Processing Models} 
\label{pcmodel}
In this section, we list the most prominent approaches for processing point-clouds. 

The \textbf{3D convolution-based method} is one of the mainstream point-cloud processing approaches which efficiently processes point-clouds based on voxelization. In this approach, voxelization is used to rasterize point-clouds into regular grids (called voxels). Then, we can apply 3D convolutions to the processed point-cloud. However, enamors empty voxels make lots of unnecessary computations. Sparse convolution is proposed to apply on the non-empty voxels~\cite{liu2015sparse,choy20194d,tang2020searching,zhou2020cylinder3d,yan2018second,feng2021simple}, largely improving the efficiency of 3D convolutions. 

The \textbf{projection-based method} attempts to project a 3D point-cloud to a 2D plane and uses 2D convolution to extract features~\cite{wang2018fusing,wu2017squeezeseg,wu2018squeezesegv2,xu2020squeezesegv3,su2015multi,lawin2017deep,boulch2017unstructured}. Specifically, bird-eye-view projection~\cite{yang2018pixor,lang2019pointpillars} and spherical projection~\cite{wu2017squeezeseg,wu2018squeezesegv2,xu2020squeezesegv3,milioto2019rangenet++} have made great progress in outdoor point-cloud tasks. 

Another approach is the \textbf{point-based method}, which directly processes the point-cloud data. The most classic methods, PointNet~\cite{qi2016pointnet} and PointNet++~\cite{qi2017pointnetplusplus}, consume points by sampling the center points, group the nearby points, and aggregate the local features. Many works further develop advanced local-feature aggregation operators that mimic the 3D convolution operation to structured data~\cite{xu2021you,li2018pointcnn,hua2018pointwise,liu2019densepoint,liu2020closer,wang2017cnn,li2018so,komarichev2019cnn}.

\subsection{Pretraining in 2D and 3D Computer Vision}
\textbf{Pretraining in 2D computer vision} is an effective approach using supervised~\cite{dosovitskiy2020image,girshick2014rich}, self-supervised~\cite{jing2020self,goyal2021self}, and contrastive learning ~\cite{he2020momentum,bachman2019learning,chen2020simple,caron2020unsupervised,chen2020improved,hjelm2018learning}. After pretraining on a large amount of data, a 2D model requires less computational and data resources for finetuning in order to obtain competitive performance on downstream tasks~\cite{kataoka2020pre,caron2019unsupervised,chen2020big,henaff2020data}.

\textbf{Pretraining in 3D computer vision} has been studied similarly as pretraining in 2D vision: both self-supervised and contrastive pretraining~\cite{xie2020pointcontrast,hou2021exploring,wang2020unsupervised} show promising results. 3D point-clouds are difficult to annotate, and there is no large-scale annotated dataset available. To address this, previous works have tried to use model pretraining to improve data efficiency \cite{xu2020weakly}. Recent works~\cite{hou2020exploring,zhang2021self} explored using contrastive learning on point-clouds. Our work does not rely on long-time pretraining. Instead, we can directly take large amounts of open-sourced image-pretrained models for a variety of point-cloud tasks.

\subsection{Cross-Modal Transfer Learning}
\textbf{Cross-modal transfer learning} takes advantage of data from various modalities~\cite{dai20183dmv,liu20213d}. For example,~\cite{liu2021learning} proposed pixel-to-point knowledge transfer (PPKT)
from 2D to 3D which uses aligned RGB and RGB-D images during pretraining. Our work does not rely on joint image-point-cloud pretraining. Instead, we directly transfer an image-pretrained model to a point-cloud model with the simplest pretraining-finetuning scheme. 

Some of the previous works for video and medical images~\cite{carreira2017quo,shan20183} have adopted the method of simply extending a pretrained 2D convolutional filter along time or depth direction for transferring to 3D models. However, the domain gaps between point-clouds and images are much more than that of videos/medical images and images. Between language and image modalities, transfer learning with minimal finetuning also shows a competitive performance~\cite{lu2021pretrained,radford2021learning}.

\subsection{Neural Collapse}
\label{nc_related}

Neural collapse (NC)~\cite{Papyan24652,han2021neural} is a recently discovered phenomenon in deep learning. It has been observed that during the training of deep overparameterized neural networks for standard classification tasks, the penultimate layer's features associated with training samples belonging to the same class concentrate around their class means. Essentially,~\cite{Papyan24652} observed that the ratio of the within-class variances and the distances between the class means converge to zero. In addition to that, it has also been observed that asymptotically the class means (centered at their global mean) are not only linearly separable, but are also maximally distant and located on a sphere centered at the origin up to scaling, and furthermore, that the behavior of the last-layer classifier (operating on the features) converges to that of the nearest-class-mean decision rule.

Recently,~\cite{galanti2022on} studied the relationship between \textbf{neural collapse and transfer learning}. They studied a transfer learning setting, where we intend to solve a target (classification) task, where only a limited amount of samples is available, so a model is pretrained and transferred from a source (classification) task. They showed that neural collapse extends beyond training and generalizes also to unseen test samples and new classes. In addition, it was shown that in the presence of neural collapse in the new classes, training a linear classifier on top of the learned penultimate layer requires only a few samples to generalize well. However, their empirical and theoretical analysis assumes that the source and target classes are i.i.d.\ samples (\textit{e.g.}, a random split of the classes in ImageNet). This implies that the two tasks share the same modality. Therefore, we suggest training an adaptor (\textit{e.g.}, a linear layer) along with retraining the normalization parameters as part of the transfer process. Intuitively, the adaptor takes samples of the second modality and translates them to representations that are interpretable by the pretrained model, such that it produces feature embeddings that are clustered into classes. In Appendix \ref{spp:theory}, we extend the framework in~\cite{galanti2022on} to the case where the source and target tasks are of different modalities and theoretically analyze it.   

\section{Converting a 2D Image Model to a 3D Point-Cloud Model}
\label{transfer}
In this paper, we primarily focus on the 3D 
sparse-convolution-based method to process point-clouds, since it can be extended to a wide range of point-cloud tasks. The other point-cloud models we use in this paper are byproducts of copying the weights of 2D image models, for example, 2D ConvNets~\cite{he2016deep} or vision transformers~\cite{dosovitskiy2020image}. In this section, we provide an in-depth introduction to how we transform the 2D ConvNets into 3D sparse ConvNets by inflation~\cite{carreira2017quo}. 

\paragraph{Inflating a 2D ConvNet into a 3D sparse ConvNet.} As discussed in Section~\ref{pcmodel}, we consider a set of points, where each point is represented by its 3D coordinates and additional features such as its intensity and RGB. We then voxelize/quantize these points into voxels according to their 3D space coordinates, following~\cite{choy20194d}. A voxel's feature is inherited from the point that lies within the voxel. If there are multiple points associated with the same voxel, we average all points' features and assign the mean to the voxel. If there is no point in the voxel, then we simply set the voxel's feature to 0. With sparse convolution, the computation on empty voxels can be skipped. 

Given a pretrained 2D ConvNet, we convert it to a 3D ConvNet that takes 3D voxels as input. The key element of this procedure is to convert 2D convolution filters to 3D, \textit{i.e.}, constructing 3D filters with the weights directly inherited from 2D filters. 
A 2D convolutional filter can be represented with a 4D tensor of shape $[M, N, K, K]$, representing output dimension, input dimension, and two spatial kernel sizes, respectively. A 3D convolutional filter has an extra dimension, and its shape is $[M, N, K, K, K]$. To better illustrate, we ignore the output and input dimensions and only consider a spatial slice of the 2D filter with shape $[K, K]$. The simplest way to convert this 2D filter to 3D is to repeat the 2D filter $K$ times along a third dimension. This operation is the same as the \textit{inflation} 
technique used by~\cite{carreira2017quo} to initialize a video model with a pretrained 2D ConvNet. 

Besides convolution, other operations such as downsampling, BN, and  nonlinear activation can be easily migrated to 3D. Our 3D model inherits the architecture of the original 2D ConvNet, but we also add a linear layer as the input layer and an output layer depending on the target task. For classification, we use a global average pooling layer followed by one fully connected layer to get the final prediction. For semantic segmentation, the output layer is a U-Net style decoder~\cite{ronneberger2015u}. The architecture of the input/output layers is described in more detail in Appendix \ref{spp:archdetail}.

\paragraph{A note on image-to-video transfer.} It is noteworthy to mention that although inflation is commonly used in video domains, image-to-point-cloud transfer is fundamentally different from image-to-video transfer. Even though videos and point-clouds are both 3D data, they are represented with completely different visual modalities with different distributions. Intrinsically, 3D point-clouds are represented as a sparse set of points lying on object surfaces and parameterized by $xyz$-coordinates, while videos are dense RGB arrays, where the two spatial arrays represent RGB images and the temporal array reflects how images evolve through time. Point-clouds are translation and rotation invariant or equi-variant, while for videos, the spatial and temporal dimensions are not interchangeable. In this paper, we surprisingly find that with simple operations such as inflation, the image-pretrained models can be directly used for point-cloud understanding under the situation that image and point-cloud are drastically different. The detailed experiments showing the feasibility and utility, and the discussion of why it works from the aspect of neural collapse are illustrated in Section \ref{exp} and Section \ref{discuss}, respectively. 

\section{Empirical Evaluation}
\label{exp}
To explore the image to point-cloud transfer, we study three settings: , {\bf (1)} finetuning input, output, and batch normalization layers (FIP-IO+BN), {\bf (2)} finetuning the whole pretrained network (FIP-ALL), and optionally {\bf (3)} partially-finetuned-image-pretrained model, only finetuning input and output layers (FIP-IO). Under the three settings, we extensively explore the feasibility of transferring the image-pretrained model for point-cloud understanding and its benefits. The entire empirical evaluation is organized as four questions: {\bf(1)} Can we transfer pretrained-image models to recognize point-clouds? (Section \ref{capable1}) {\bf (2)} Can image-pretraining benefit the performance of point-cloud recognition? (Section \ref{performance}) {\bf (3)} Can image-pretrained models improve the data efficiency on point-cloud recognition? (Section \ref{efficiency}) {\bf(4)} Can image-pretrained models accelerate training point-cloud models? (Section \ref{speed}) 

\paragraph{Datasets.} We evaluate the transferred models on ModelNet 3D Warehouse classification~\cite{Wu_2015_CVPR}, S3DIS indoor segmentation~\cite{armeni2017joint}, and SemanticKITTI outdoor segmentation~\cite{behley2019iccv} tasks. ModelNet 3D Warehouse is a CAD model classification dataset that consists of point-clouds with 40 categories. CAD models in this benchmark come from 3D Warehouse~\cite{3dwarehouse}. In this benchmark, we only utilize $x, y, z$ coordinates as features. S3DIS is a dataset collected from real-world indoor scenes and includes 3D scans of Matterport Scanners from 6 areas. It provides point-wise annotations for indoor objects like chair, table, and bookshelf, \textit{etc.} SemanticKITTI dataset from KITTI Vision Odometry~\cite{geiger2012cvpr} is a driving scene dataset. It provides dense point-wise annotations for the complete 360 degrees field-of-view of the deployed automotive lidar, which is currently one of the most challenging datasets. 

ResNet~\cite{he2016deep} series is used mostly throughout our experiments. Depending on the experiments, ResNets are pretrained on Tiny-ImageNet, ImageNet-1K, ImageNet-21K~\cite{deng2009imagenet}, and Fractal database (FractalDB)~\cite{kataoka2020pre}. Our pretrained models are directly downloaded from various sources, with detailed links provided in the Appendix \ref{spp:impdetail}. To study the benefits of using pretrained image models, we also utilize PointNet++~\cite{qi2017pointnetplusplus}, ViT~\cite{dosovitskiy2020image}, and SimpleView~\cite{goyal2021revisiting} as our baselines.  

\subsection{Can we transfer pretrained-image models to recognize point-clouds?}
\label{capable1}

To evaluate the feasibility of transferring pretrained 2D image models to 3D point-cloud tasks, we conduct experiments on top of the ResNet series since there are abundant open-source pretrained ResNet available. In particular, we convert 2D ConvNets into 3D ConvNets using the procedure described in Section \ref{transfer}. We hypothesize that, if a pretrained 2D image model is capable of understanding point-clouds directly, we can see a non-trivial performance by only finetuning input and output layers of the transferred model. Further, as we gradually relax the frozen parameters, finetuning BN parameters as well, the transferred model can achieve better performance, even surpassing training-from-scratch. 

\begin{figure*}[!t]
    \centering
    \includegraphics[width=1.0\textwidth]{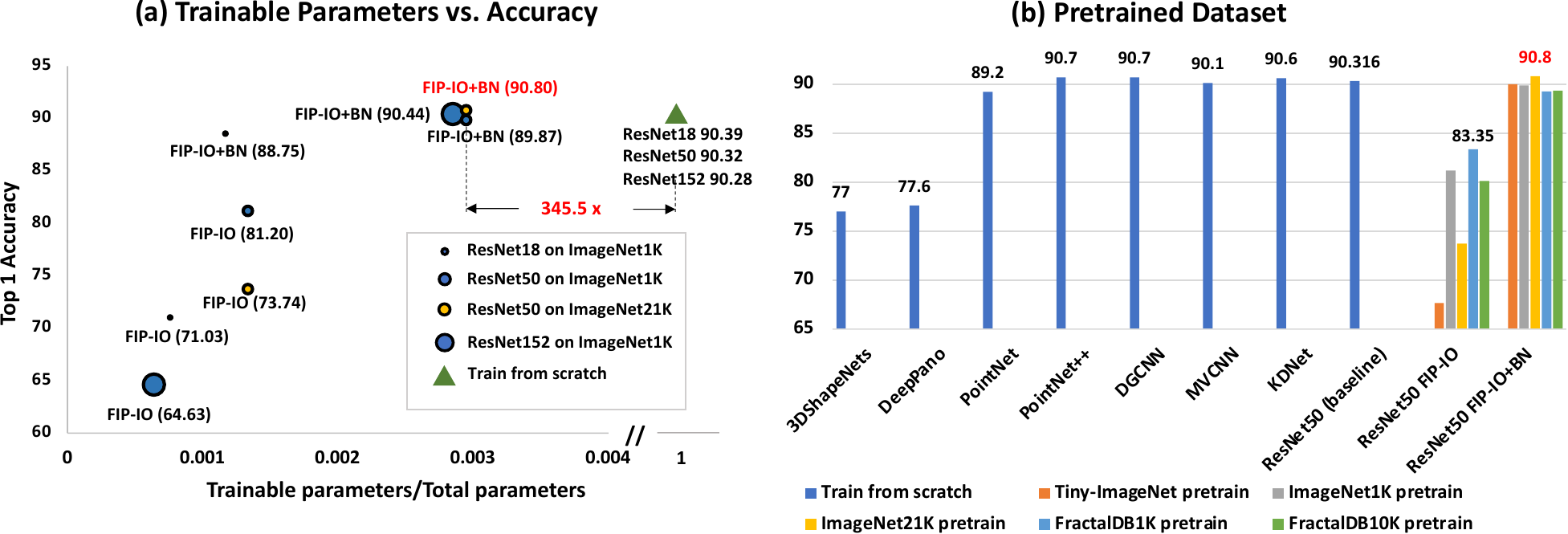}
    \caption{\textbf{a)} the left figure shows the trainable parameters ratio \textit{w.r.t} top-1 accuracy on ModelNet 3D Warehouse dataset.  \textbf{b)} the right figure shows the performance of FIP-IO and FIP-IO+BN on top of \textbf{ResNet50} pretrained on different datasets. }
    \label{fig:exp_pfip}
\end{figure*}

We conduct two groups of experiments with FIP-IO and FIP-IO+BN, with the results shown in Figure \ref{fig:exp_pfip}. The first is to evaluate the performance as the trainable parameters gradually increase. As shown in Figure \ref{fig:exp_pfip} (a), training \textbf{no more than 0.3 \% (345.5x fewer)} of the whole parameters, the image pretraining even beats the training-from-scratch (100 \% trainable parameters). Specifically, ResNet152 FIP-IO+BN with ImageNet1K pretraining improves training-from-scratch by 0.16 points, and ResNet50 FIP-IO+BN with ImageNet21K pretraining improves 0.48 points. Meanwhile, FIP-IO reaches a non-trivial performance. ResNet50 FIP-IO pretrained on ImageNet1K achieves 81.20 \% top-1 accuracy, only 9.12 points worse than training-from-scratch with approximately 0.1 \% trainable parameters. 

Furthermore, to investigate the effect of different datasets, as shown in the right figure of Figure \ref{fig:exp_pfip}, we inflate ResNet50 pretrained from different image datasets, including Tiny-ImageNet, ImageNet1K, ImageNet21K, FractalDB1K, and FractalDB10K, then evaluate on the ModelNet 3D Warehouse. 

We discover that, even if we only finetune the input and output layers while keeping the image-pretrained weights frozen, the FIP-IO pretrained from ImageNet1K, FractalDB1K, and FractalDB10K achieves competitive performance. Specifically, ResNet50 FIP-IO with ImageNet1K pretraining outperforms 3D ShapeNet~\cite{Wu_2015_CVPR} and DeepPano~\cite{7273863}, which were the state-of-the-arts in 2015, by 4.2 and 3.6 points respectively in top-1 accuracy on ModelNet 3D Warehouse. More importantly, with ImageNet21K pretrained model, ResNet50 FIP-IO+BN surpasses training-from-scratch by 0.48 points, even beating a variety of well-known methods including PointNet~\cite{qi2016pointnet}, MVCNN~\cite{su2015multi}, DGCNN~\cite{wang2019dynamic}, \textit{etc}. 

Notably, we find out the answer to "Can we transfer pretrained-image models to recognize point-clouds?": Yes. The pretrained 2D image models can be directly used for recognizing point-clouds. Surprisingly, the pretraining dataset is not restricted to natural but also synthetic images like those in FractalDB1K/10K.

\begin{table}[!t]
    \centering
    \setlength{\tabcolsep}{1.1mm}{
    \footnotesize
    \caption{ModelNet 3D Warehouse classification results (top-1 accuracy \%) of fully-finetuned-image-pretrained models (FIP-ALL) based on different pretrained models. We include 2021 SOTAs, such as RSMix~\cite{lee2021regularization}, Point Transformer (Point-Trans)~\cite{zhao2021point}, DRNet~\cite{qiu2021dense}, and PointCutMix~\cite{zhang2021pointcutmix},  for comparison.}

    \begin{tabular}{ p{35mm}<{\centering}| p{22mm}<{\centering}  p{22mm}<{\centering}  p{22mm}<{\centering}  p{22mm}<{\centering}}
    \toprule[1pt]
    \textbf{Method} &  \textbf{ResNet18} & \textbf{ResNet50} & \textbf{ResNet152} & \textbf{ResNet101$\times$2}\\
    \midrule[0.5pt]
    From Scratch & 90.39 & 90.32 & 90.28 & 90.03   \\
    FIP-ALL on ImageNet1K & 90.52 \textbf{(+0.13)} & 90.92 \textbf{(+0.60)} & 91.09 \textbf{(+0.81)} & 90.52 \textbf{(+0.49)} \\
    FIP-ALL on ImageNet21K & - & 91.05 \textbf{(+0.73)} & -  & -\\
    \bottomrule[1pt]
    \end{tabular}
    
    ~\\
    
    \begin{tabular}{ p{35mm}<{\centering}| p{22mm}<{\centering}  p{22mm}<{\centering}  p{22mm}<{\centering}  p{22mm}<{\centering}}
    \toprule[1pt]
    \textbf{Method} & \textbf{PointNet++(SSG)} & \textbf{ViT-B-16} & \textbf{ViT-L-16} & \textbf{SimpleView}\\
    \midrule[0.5pt]
    From Scratch &90.34  & 84.27 & 83.48 & 93.3  \\
    FIP-ALL on ImageNet1K&91.22 \textbf{(+0.88)} & - &-  & 93.8 (\textbf{+0.50}) \\
   FIP-ALL on ImageNet21K & -& 87.77 \textbf{(+3.50)} & 87.66 \textbf{(+4.18)} & -\\
    \bottomrule[1pt]
    \end{tabular}
    
    ~\\
    
    \begin{tabular}{ p{35mm}<{\centering}| p{22mm}<{\centering}  p{22mm}<{\centering}  p{22mm}<{\centering}  p{22mm}<{\centering}}
    \toprule[1pt]
    \textbf{Method} & \textbf{RSMix} & \textbf{Point-Trans} & \textbf{DRNet} & \textbf{PointCutMix}\\
    \midrule[0.5pt]
    From Scratch &93.5  & 93.7 & 93.1 & 93.4  \\
    \bottomrule[1pt]
    \end{tabular}
    \label{tab:performance1} 
    }
\end{table}

\begin{table}[!t]
\footnotesize
\caption{Indoor scene and outdoor scene segmentation results (mIoU \%) of fully-finetuned-image-pretrained Model (FIP-ALL). In this table, all image-pretrained models are pretrained on ImageNet1K.}
    \begin{tabular}{ c| c c | c c }
    \toprule[1pt]
    \multirow{2}*{\textbf{Method}} & \multicolumn{2}{c|}{\textbf{S3DIS (mIoU \%)}} & \multicolumn{2}{c}{\textbf{SemanticKITTI (mIoU \%)}} \\
    \cmidrule{2-5}
    ~ & \textbf{PointNet++(SSG)} & \textbf{ResNet18} & \textbf{HRNetV2-W48} & \textbf{ResNet18}\\
    \midrule[0.5pt]
    From Scratch &  52.45 & 55.09 & 44.12 &  64.75 \\
    FIP-ALL on ImageNet1K & 55.01 (\textbf{+2.56}) & 56.62 (\textbf{+1.53}) & 47.53 (\textbf{+3.41}) & 65.57 (\textbf{+0.82})   \\
    % \midrule[0.1pt]
    \bottomrule[1pt]
    \end{tabular}
\label{tab:performance2}
\end{table}

 \begin{table}[!t]
    \centering
    \footnotesize
    \caption{Comparison with PointContrast~\cite{xie2020pointcontrast} on the ModelNet 3D Warehouse. PointContrast provides two different pretrained models with using PointInfoNCE loss and Hardest Contrastive loss, respectively.}
    \begin{tabular}{ p{22mm}<{\centering}| p{25mm}<{\centering}  p{28mm}<{\centering} | p{40mm}<{\centering}}
    \toprule[1pt]
   \textbf{From scratch} & \textbf{PointInfoNCE} &  \textbf{Hardest Contrastive} & \textbf{ImageNet1K pretrain (Ours}) \\
  \midrule[0.5pt]      
89.95 & 90.24 (\textbf{+0.29}) & 90.15 (\textbf{+0.20})& 90.88 (\textbf{+0.93})\\
    \bottomrule[1pt]
    \end{tabular}
    \label{tab:pointcontrast}
    \end{table}

\subsection{Can image-pretraining benefit point-cloud recognition?}
\label{performance}

From the previous subsection, we find unexpectedly that the image-pretrained model can be directly used for point-cloud understanding. In this subsection, we investigate whether the image-pretrained model is helpful to improve the performance of point-cloud tasks. We use different baselines, including voxelization-based method (simply ResNet), point-based method (PointNet++~\cite{qi2017pointnetplusplus}), projection-based method (SimpleView~\cite{goyal2021revisiting}), and current popular transformer-based method (ViT-B-16 and ViT-L-16~\cite{dosovitskiy2020image}), and fully finetune them on three point-cloud datasets: classification on ModelNet 3D Warehouse, indoor scene segmentation on S3DIS, and outdoor scene segmentation on SemanticKITTI, as shown in Table \ref{tab:performance1} and Table \ref{tab:performance2}.

For PointNet++, we use ImageNet1K to pretrain: we break each image into pixels and regard it as a point-cloud. For ViT, we directly use the open-source pretrained model and finetune it on ModelNet 3D Warehouse. All the implementation details are illustrated in Appendix \ref{spp:impdetail}.

Table \ref{tab:performance1} presents performance on ModelNet 3D Warehouse dataset. We observe that FIP-ALL improves all baselines steadily and significantly. Besides, pretraining brings more improvements to deeper models. For example, ResNet18 can only be improved by 0.13\% top-1 accuracy, but pretraining on ImageNet1K leads to 0.81 points top-1 accuracy improvement on top of ResNet152. Moreover, larger pretrained datasets also lead to better performance. Specifically, ResNet50 FIP-ALL from ImageNet21K can reach 91.05\% top-1 acc, with 0.73 points improvement over training-from-scratch. Such FIP-ALL significantly outperforms a series of well-known methods such as~\cite{qi2016pointnet,qi2017pointnetplusplus,klokov2017escape,wang2019dynamic,su2015multi,li2018so}.

We also explore FIP-ALL on different architectures, as shown in the second group of Table \ref{tab:performance1}. In particular, FIP-ALL on top of PointNet++, ViT-B-16, ViT-L-16, and SimpleView with image dataset pretraining improve the training-from-scratch by 0.88, 3.50, 4.18, 0.50 points, respectively. Especially for the current superior baseline in image recognition, ViT-B-16 and ViT-L-16, the improved performance is quite significant, revealing the huge potential of using image-pretrained models for point cloud recognition.

For the challenging indoor and outdoor scene segmentation, using ImageNet1K pretrained models (FIP-ALL on ImageNet1K) also improve the training-from-scratch consistently, as shown in Table \ref{tab:performance2}. PointNet++ (resp. ResNet18) pretrained on ImageNet1K outperforms the training-from-scratch by 2.56 points (resp. 1.53 points) mIoU on S3DIS dataset. For SemanticKITTI, we utilize the commonly used projection-based method with 2D ConvNet HRNet. With ImageNet1K pretraining, we observe 3.41 points mIoU improvement, a large margin in such a challenging task. Since HRNetV2-W48 has rich pretrained models, we finetune Cityscapes pretrained HRNetV2-W48 and observe this enhances more (5.25\% mIoU improvement over training from scratch). Even for the ResNet18 with a high from-scratch performance of 64.75\% mIoU, the ImageNet1K pretraining can also bring 0.82 points mIoU improvement.

Finally, we compare the performance gain with the well-known point-cloud self-supervised method PointContrast~\cite{xie2020pointcontrast}, as presented in Table \ref{tab:pointcontrast}. We use the same model architecture and finetuning recipe, and the only difference is the pretraining weights. Note that the model architecture used in PointContrast does not have corresponding open-sourced image-pretrained weights, so we pretrain it by ourselves on ImageNet1K, with the standard ImageNet training recipe provided by Pytorch. We can observe that image-pretraining on ImageNet1K significantly boosts the training-from-scratch by 0.93 points, surpassing the PointContrast by at least 0.64 points. 

Therefore, the answer to "Can image-pretraining benefit point-cloud recognition" is: Yes. Image-pretraining can indeed improve point-cloud recognition, which can generalize to a wide range of backbones and benefit a variety of challenging tasks.

\subsection{Can image-pretrained models improve the data efficiency on point-cloud recognition?}
\label{efficiency}

\begin{table}[!t]
    \centering
    \setlength{\tabcolsep}{1.1mm}{
    \footnotesize
    \caption{Few-shot experiments on top of different ResNets on the ModelNet 3D Warehouse dataset. We conduct 3 trials for each setting and results are as mean $\pm$ std.}
    \begin{tabular}{ m{14mm}<{\centering}| m{37mm}<{\centering}  m{37mm}<{\centering}  m{37mm}<{\centering} }
    \toprule[1pt]
    \multirow{2}*{\textbf{Few-shot}}
    %\textbf{Few shot (from scratch \& FIP-ALL)} 
    & \textbf{ResNet18} & \textbf{ResNet50} & \textbf{ResNet152}\\
    ~&\multicolumn{3}{c}{\textbf{(from scratch/FIP-ALL)}}\\
    \midrule[0.5pt]
    10-shot & 72.2$\pm$0.8/73.2$\pm$0.6 \textbf{(+1.0)} &
    71.7$\pm$0.7/74.1$\pm$0.8 \textbf{(+2.4)}& 
    69.8$\pm$1.1/73.9$\pm$0.4  \textbf{(+4.1)} \\
    5-shot & 63.7$\pm$1.6/66.6$\pm$0.8 \textbf{(+2.9)}& 62.4$\pm$1.1/66.0$\pm$2.2 \textbf{(+3.6)} & 
    59.4$\pm$0.8/66.5$\pm$0.9 \textbf{(+7.1)} \\
    1-shot & 26.8$\pm$4.4/36.8$\pm$0.6 \textbf{(+10.0)} & 28.1$\pm$0.4/34.1$\pm$0.2 \textbf{(+6.0)} & 
    23.3$\pm$4.3/33.2$\pm$1.3 \textbf{(+9.9)} \\
    \bottomrule[1pt]
    \end{tabular}
    \label{tab:fewshot}
    }
\end{table}

\begin{table}[!t]
    \centering
    \setlength{\tabcolsep}{1.1mm}{
    \footnotesize
    \caption{Semi-supervised distillation experiments on top of ResNet34 on the ModelNet 3D Warehouse dataset.}
    \begin{tabular}{ m{14mm}<{\centering}| m{25mm}<{\centering}  m{25mm}<{\centering}  m{25mm}<{\centering} m{25mm}<{\centering}}
    \toprule[1pt]
    \textbf{Few-shot}
    & \textbf{From scratch} & \textbf{PointInfoNCE} & \textbf{Hardest Contrastive} & \textbf{ImageNet1K pretrain (Ours)}\\
    \midrule[0.5pt]
    10-shot & 72.2 & 74.6 \textbf{(+2.4)} & 74.6 \textbf{(+2.4)} & 74.9 \textbf{(+2.7)}\\
    5-shot & 61.9 & 65.1 \textbf{(+3.2)} & 65.9 \textbf{(+4.0)} & 66.0 \textbf{(+4.1)}\\
    1-shot & 29.2 & 39.0 \textbf{(+9.8)} & 37.2 \textbf{(+8.0)} & 41.1 \textbf{(+11.9)}\\
    \bottomrule[1pt]
    \end{tabular}
    \label{semi}
    }
\end{table}

Data efficiency is extremely important in point-cloud understanding due to the huge labor of collecting and annotating point-cloud data. In this subsection, we investigate whether the image-pretrained model can help to improve the data efficiency by conducting few-shot setting experiments, including 1-shot, 5-shot, and 10-shot. 

In detail, for each class (ModelNet 3D Warehouse involves 40 classes), we randomly choose a few point-clouds as training data and still evaluate on the whole test set. We compare the results between training-from-scratch and FIP-ALL pretrained on the ImageNet1K dataset. The experimental results are shown in Table \ref{tab:fewshot}. We observe that FIP-ALL dramatically surpasses training-from-scratch on the low data regime (1-shot): pretraining on ImageNet1K brings 10.0, 6.0, and 9.9 points top-1 accuracy improvement for ResNet18, ResNet50, and ResNet152, respectively. For 5-shot and 10-shot settings, using ImageNet1K pretraining can still consistently improve the performance.

Furthermore, inspired by previous work \cite{chen2020big} which proposed \textit{big self-supervised models are strong semi-supervised learners} in 2D image recognition, we borrow the idea and propose \textit{an image-pretrained model is also a strong semi-supervised learner in point-cloud recognition}. We also compared the image-pretrained model with the self-supervised pretrained model in this experiment. Specifically, we first take pretrained models from the previous self-supervised pretraining method PointContrast \cite{xie2020pointcontrast}. PointContrast provides two ScanNet \cite{dai2017scannet} pretrained models of architecture ResNet34 trained with hardest-contrastive loss and PointInfoNCE loss. Then, we finetune PointContrast on 1/5/10 shot of the labeled ModelNet 3D Warehouse dataset and regard it as a teacher model. Finally, we distill the teacher model to a randomly initialized student model. In detail, we pass in the rest of unlabeled ModelNet 3D Warehouse dataset and 1/5/10 shot of the labeled dataset into the teacher model to generate pseudo labels. We use softmax MSE loss as consistency loss between student model outputs and pseudo labels. When the data instance is labeled, we add an additional cross entropy loss as a class criterion between student output and the label. 

To show the effectiveness of the image-pretrained model, we repeat the above experiment, only replacing self-supervised pretrained models with ResNet34 ImageNet1K pretrained models. Results are reported in Table \ref{semi}. We observe that image-pretrained ResNet34 consistently outperforms PointContrast, and improves the baseline by a large margin with 11.9, 4.1, and 2.7 points on 1-shot, 5-shot, and 10-shot, respectively. The results in Table \ref{semi} show that an image-pretrained model is indeed a strong semi-supervised learner in point-cloud recognition. 

However, in both Table \ref{tab:fewshot} and Table \ref{semi}, we observe that as the amount of training data increases, the performance gain becomes saturated. Therefore, our answer to "Can image-pretrained models improve the data efficiency on point-cloud recognition?" is: Yes. Image-pretrained models can improve the data efficiency on point-cloud recognition, especially on low data regime. When the training data increases, performance still improves, but the gain becomes marginal.

\begin{figure*}[!t]
    \centering
    \includegraphics[width=1.0\textwidth]{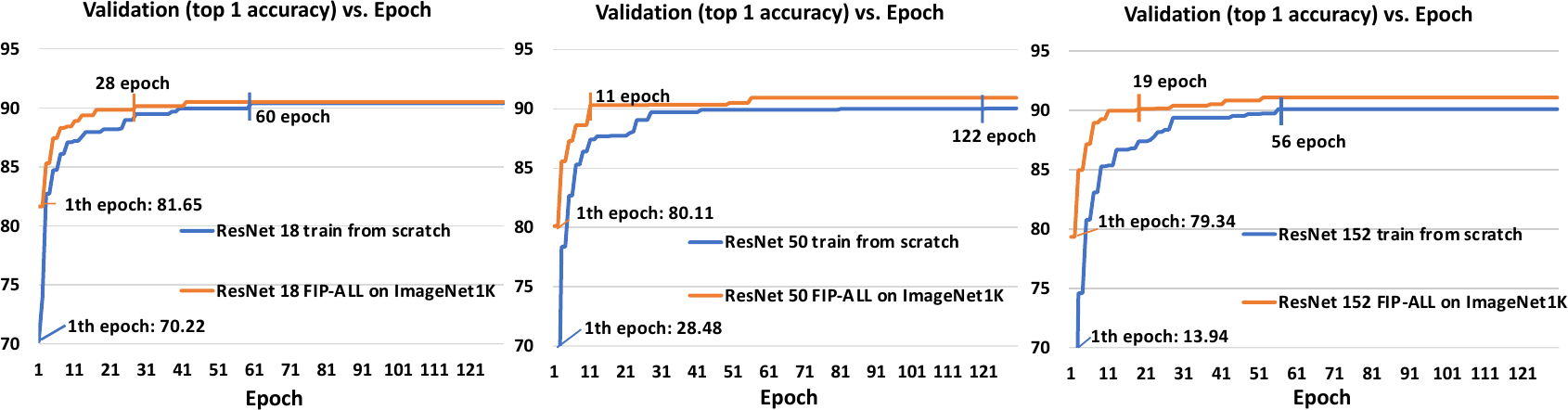}
    \caption{{\bf Comparing the validation accuracy of training-from-scratch and FIP-ALL on ModelNet 3D.} We report the validation accuracy during training. We compare the results between training-from-scratch and fine-tuning the whole network (FIP-ALL) with pretraining ResNet18, ResNet50, and ResNet152, on the ImageNet1K dataset.}
    \label{fig:trainspeed}
\end{figure*}

\subsection{Can image-pretrained models accelerate point-cloud training?}
\label{speed}
We also investigate whether the image-pretrained model can accelerate training on the point-cloud domains. The results are shown in Fig.~\ref{fig:trainspeed}.

We discover that, after training only one epoch on ModelNet 3D Warehouse dataset, FIP-ALL pretrained on ImageNet1K achieves very impressive performance, yet the performance of training-from-scratch is still very low. For instance, after the first epoch, ResNet50 (resp. ResNet152) with training from scratch achieves 28.48\% (resp. 13.94\%) top-1 accuracy while ResNet50 (resp. ResNet152) with ImageNet1K pretraining reaches 80.11\% (resp. 79.34\%) top-1 accuracy. Moreover, to reach 90\% top-1 accuracy, a non-trivial performance, FIP-ALL significantly accelerates the training by 2.14x (28 vs. 60 epoch), 11.1x (11 vs. 122 epoch), 2.95x (19 vs. 56 epoch) over training-from-scratch, on top of ResNet18, ResNet50, and ResNet152, respectively.

Therefore, our answer to ``Can image-pretrained models accelerate point-cloud training?'' is still positive. The image-pretrained models can significantly accelerate the training speed of point-cloud tasks.

\begin{figure*}[!t]
    \centering
    \includegraphics[width=.95\textwidth]{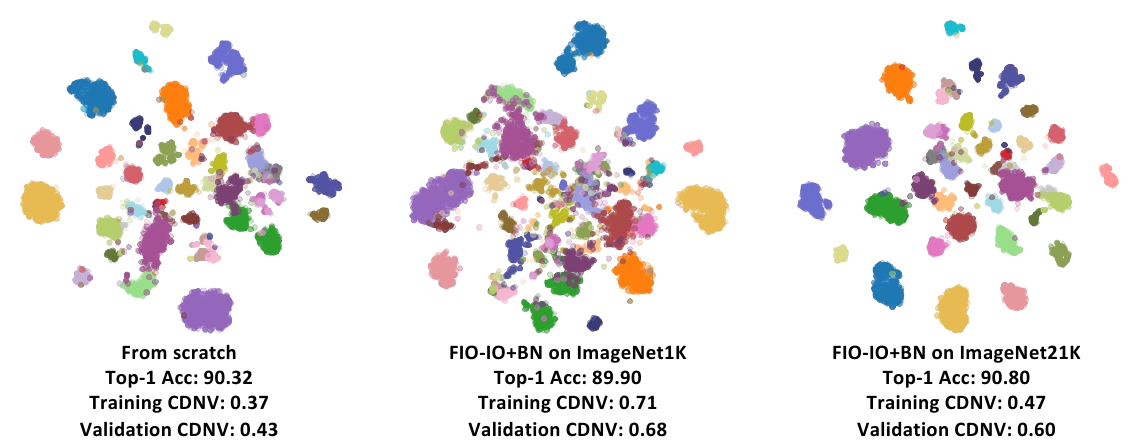}
\caption{tSNE visualization and class-distance normalized variance on ModelNet 3D Wharehouse dataset. FIP-IO+BN on ImageNet1K/21K are the same models in Fig.~\ref{fig:exp_pfip}. CDNV is computed for the fine-tuned model on the train and test data of the point-cloud domain.}
    \label{fig:tsne}
\end{figure*}

\section{Neural Collapse in Cross-Modal Transfer}
\label{discuss}
In this section, we provide an explanation of why the image to point-cloud transfer works based on the recently observed phenomenon called neural collapse \cite{han2021neural,Papyan24652}. \cite{galanti2022on} in depth studied the relationship between neural collapse and transfer learning between two classification tasks of the same modality (image domain). Similar to this work, we focus on transferring pretrained models between domains of different modalities, \textit{i.e.}, from images to point-clouds.

As illustrated in Section~\ref{exp}, we can transfer models that were pretrained on images to the point-cloud domain. This motivates us to question whether the phenomenon of neural collapse generalization~\cite{galanti2022on} (see Section~\ref{nc_related}) is also evident in our case. Following~\cite{galanti2022on}, we explore the relationships between neural collapse and image-to-point transfer by calculating the class-distance normalized variance (CDNV). Informally, the CDNV measures the ratio between the within-class variances of the embeddings and the squared distance of their means (see Appendix \ref{sec:nc} for details). We measure the CDNV of the fine-tuned model on both train and test data of the point-cloud domain. Since neural collapse is essentially a clustering property of features learned by neural networks, we further examine the neural collapse using tSNE visualizations. The results are summarized in Fig.~\ref{fig:tsne}.

We observe that with finetuning much fewer (345.5x fewer) parameters in ResNet50 pretrained on ImageNet1K, both class-distance-normalized-variance and the clustering of tSNE are worse than training-from scratch, but still show relatively obvious clustering phenomenon. However, when we use the ResNet50 pretrained on ImageNet21K, the top-1 accuracy, and CDNV are significantly improved. More importantly, CDNV of ImageNet1K pretrained ResNet50 and ImageNet21K pretrained ResNet50 is lower than 1. This observation indicates although the image domain and point-cloud domain are quite different, the phenomenon of neural collapse generalization~\cite{galanti2022on} still exists in their transfer. More results and analysis are illustrated in Appendix \ref{sec:nc}.

Moreover, the interesting discovery pushes us to think about the reason of cross-modal transfer having neural collapse. Inspired by \cite{galanti2022on}, we briefly explain below. More detailed theoretical proof is presented in Appendix \ref{spp:theory}.

\paragraph{Theoretical idea.} In this work we focused on the problem of transferring knowledge between two tasks (source and target) consisting of two different modalities with different classes. Therefore, in the theoretical analysis, we have to deal with two separate modes of generalization: \textit{between classes} and \textit{between modalities}. In order to model this problem, we assume that the target and source tasks are decomposed of i.i.d.\ classes that are samples of two different distributions $\cD_1$ and $\cD_2$ (each stands for a different domain/modality). Each class is defined by a distribution over samples (e.g., samples of dog images). Given a target task (consisting of a set of randomly selected classes $P_1,\dots,P_k \sim \cD_1$), the pretrained model is evaluated after training an adaptor and a linear classifier on top of it. Its overall performance is measured in expectation over the selection of target tasks. 

To capture the similarity between the two domains, we assume there exists an invertible mapping $F$ between the classes that preserves the density of the two distributions, namely, $\hP_c = F(P_c)\sim \cD_2$ for $P_c \sim \cD_1$. To characterize the similarity between the classes coming from $\cD_1$ and $\cD_2$, we further assume that the classes $P_c$ and $\hP_c$ share a `mutual representation space' from which the class label can be recovered. The shared space is given by two simple functions $g^*$ and $\tig^*$ for which the distance between $g^* \circ P_c$ and $\tig^* \circ \hP_c$ is small (in expectation over $P_c \sim \cD_1$). By utilizing tools from the theory of Unsupervised Domain Adaptation~\cite{bendavid,DBLP:conf/alt/Mansour09,DBLP:conf/colt/MansourMR09}, we translate the performance of a pretrained model on randomly selected target tasks into its expected error on randomly selected tasks with classes from $\cD_2$. Then, in order to bound this error, we use Proposition 5 in~\cite{galanti2022on} that relates the error and the degree of neural collapse of the pretrained model on randomly selected classes from $\cD_2$. Finally, according to Propositions 1 and 2 in~\cite{galanti2022on}, this quantity can be upper bounded by the degree of neural collapse of the pretrained model on the source train data.

\section{Conclusions}

In this work, we use finetuned-image-pretrained models (FIP) to explore the feasibility of transferring image-pretrained models for point-cloud understanding and the benefits of using image-pretrained models on point-cloud tasks. We surprisingly discover that, with simply transforming a 2D pretrained ConvNet and minimal finetuning --- input, output, and batch normalization layer (FIP-IO or FIP-IO+BN), FIP can achieve very competitive performance on 3D point-cloud classification, beating a wide range of point-cloud models that adopt a variety of tricks. Moreover, we find that when finetuning all the parameters of the pretrained models (FIP-ALL), the performance can be significantly improved on point-cloud classification, indoor and outdoor scene segmentation. Fully finetuned models generalize to most of the popular point-cloud methods. We also find that FIP-ALL can improve the data efficiency on few-shot learning and accelerate the training speed by a large margin. Additionally, we explore the relationships between neural collapse and cross modal transferring for our case, and shed light on why it works based on neural collapse. Compared with previous works that seek improvements from designing architectures and pretraining only on point-cloud modality, our work is not limited by the architecture design and the small-scale point-cloud dataset. We believe that image pretraining is one of the solutions to the bottleneck of point-cloud understanding and hope this direction can inspire the research community.

\section*{Acknowledgements}
Co-authors from UC Berkeley were sponsored by Berkeley Deep Drive (BDD). Tomer Galanti's contribution was supported by the Center for Minds, Brains and Machines (CBMM), funded by NSF STC award CCF-1231216. 

{
\normalem
\bibliographystyle{unsrt}
\bibliography{ref}
}

%%%%%%%%%%%%%%%%%%%%%%%%%%%%%%%%%%%%%%%%%%%%%%%%%%%%%%%%%%%%
%%%%%%%%%%%%%%%%%%%%%%%%%%%%%%%%%%%%%%%%%%%%%%%%%%%%%%%%%%%%

\newpage
\appendix
\section{Implementation Details}
\label{spp:impdetail}

\paragraph{Datasets.} Our experiments are conducted on ModelNet 3D Warehouse, S3DIS, and SemanticKITTI datasets. For the ModelNet 3D Warehouse dataset, we train all models on the train set and evaluate on the validation set. For the S3DIS, we train all models on area 1, 2, 3, 4, 6 and evaluate on area 5. For the SemanticKITTI dataset, we train all models on splits 00-10 except 08 which is used for evaluation. For each of the datasets, all ResNet series models use the same training scheme, and all experiments are implemented with PyTorch.

\paragraph{Training on ModelNet 3D Warehouse dataset.} In this case, coordinates of point-clouds are randomly scaled, translated, and jittered. We employ the SGD optimizer with momentum 0.9, weight-decay $10^{-4}$, and initial learning rate 0.1 with cosine learning rate scheduler. Each mini batch is set to 32, and models are trained for 300 epochs. For both training and inference phase, we only utilize $x, y, z$ coordinates without other features and set the voxel size to be 0.05. The experiments for ModelNet 3D Warehouse are all conducted on a Titan RTX GPU. 

\paragraph{Training on the S3DIS dataset.} In this case, we concatenate all subparts of an indoor scene to train and validate on. Along $x, y$ directions, scenes are applied horizontal flip randomly. RGB features are randomly jittered, translated, and auto contrasted. Finally, we normalize and clip point-clouds. We set voxel size to 0.05, use SGD optimizer with momentum 0.9, weight-decay $10^{-4}$, and initialize learning rate to 0.1 with polynomial learning rate scheduler. Each mini batch is set to 3, and models are trained for 400 epochs on 2 Titan RTX GPUs. 

\paragraph{Training on the SemanticKITTI dataset.} In this case, coordinates of each point-cloud are randomly scaled and rotated. We use SGD optimizer with momentum 0.9, weight-decay $10^{-4}$, and initial learning rate 0.24 with cosine warmup learning rate scheduler. Each mini batch is set to 2, and models are trained for 15 epochs on 4 Titan RTX GPUs. For both training and inference phases, we utilize $x, y, z$ coordinates as well as intensity feature and set voxel size to 0.05.

Most of our pretrained models were taken from open-sources~\footnote{https://pytorch.org/vision/stable/models.html}\footnote{https://github.com/Alibaba-MIIL/ImageNet21K}\footnote{https://github.com/hirokatsukataoka16/FractalDB-Pretrained-ResNet-PyTorch}\footnote{https://github.com/HRNet/HRNet-Semantic-Segmentation/tree/pytorch-v1.1}\footnote{https://github.com/rgeirhos/Stylized-ImageNet}\footnote{https://github.com/wielandbrendel/bag-of-local-features-models}, so we do not need to take time and computational resources for pretraining. We use torchsparse\footnote{https://github.com/mit-han-lab/torchsparse} to produce sparse 3D convolutions.

\paragraph{Details on Section \ref{capable1}} 
In this section, we take the ResNet architecture, inflate the pretrained models of different image datasets, and add linear input and output layers as shown in Section~\ref{spp:archdetail}. The ResNet50 was pretrained on ImageNet1K and is taken from the original PyTorch example. We use the same training recipe provided by PyTorch to train the ResNet50 on Tiny-ImageNet. The pretrained ResNet50 on ImageNet21K was taken from~\cite{ridnik2021imagenet21k}. 

\paragraph{Details on Sections \ref{performance}, \ref{efficiency}, and \ref{speed}.} In these sections, the pretrained ResNet models are taken from the same sources as those in Section \ref{capable1}.

For pretraining PointNet++ on ImageNet1K, we utilize the PointNet++ SSG version~\cite{qi2017pointnetplusplus}. We break the image into pixels and regard the group of pixels as a point-cloud with coordinates of $x, y$ positions in the original image and appending $z=1$ to all pixels. Then, we set center sampling number to 1024 and 256 for first and second stage, and the radius is set into 8 and 64, respectively. For each center point, we query 64 neighboring points. The training recipe is also provided by PyTorch. 

For ViT models, we directly take the pretrained weights from~\cite{dosovitskiy2020image}. To apply it on ModelNet 3D Warehouse, we sample 256 centers and group 64 nearby points, regarding these as ``point-cloud patches''. Then, we use a linear embedding to project the point-cloud patches into a sequence, and ViT processes them the same as image patches. Except for the linear embedding and the final output classifier, all the models are kept the same as the original version. For the experiments on S3DIS and SemanticKITTI, the architectural detail of ResNet18 is shown in A.4 listing \ref{main_segarch}. 

For SimpleView model, all the experiment settings are the same as~\cite{goyal2021revisiting}. The only difference is whether to use the pretrained ResNet18. For HRNetV2-W48, we directly use the ImageNet1K and Cityscape pretrained models from~\cite{SunXLW19}. 

We conduct three trials on the few-shot experiments. For each trial, we change the random seed but keep all the other settings the same. To plot the training speed curve, we directly use the training log without any other changes, such as smoothing. 

\section{Additional Experiments}

\subsection{Finetuning the mean and variance of batch normalization.}
For the first group of experiments, ResNet50 FIP either has IO or IO+BN finetuned. In addition to these two experimental settings, we also investigate finetuning input, output layers, and mean, variance of normalization layers, while fixing the convolution layer weights, normalization layer weights, and bias. The full experiment results with this extra setting are reported in Table \ref{tab:different_ms} and \ref{tab:depth_ms}. We can observe that compared with only finetuning input and output layers, updating mean and variance can also largely improve the performance of point-cloud recognition. As suggested in Section \ref{nc_related}, we train batch normalizations to enhance the adaption between modalities.

\begin{table}[!t]

\centering
    \centering
    \setlength{\tabcolsep}{1.1mm}{
    \footnotesize
    \caption{ResNet50 results (evaluated on ModelNet 3D Warehouse) of finetuning the mean and variance in batch normalization layers (BN) on different image-pretrained-models. IO (FIP-IO) indicates finetuning input and output layer. IOms indicates updating input and output layer, BN mean and variance. IOmsWb (FIP-IO+BN) indicates finetuning input, output layer, and the whole BN. }
    \resizebox{\linewidth}{!}{
    \begin{tabular}{ m{20mm}<{\centering}| m{20mm}<{\centering}  m{20mm}<{\centering}  m{20mm}<{\centering} m{20mm}<{\centering} m{20mm}<{\centering} }
    \toprule[1pt]
    \textbf{Layers} &  \textbf{Tiny-ImageNet} & \textbf{ImageNet1K} & \textbf{ImageNet21K} & \textbf{FractalDB1K} & \textbf{FractalDB10K} \\
    \midrule[0.5pt]
    IO & 67.67 & 81.20 & 73.74 & 83.35 & 80.11 \\
    IOms & 83.79 & 82.94 & 84.08 & 72.33 & 79.66 \\
    IOmsWb & 89.99 & 89.87 & 90.80 & 89.26 & 89.34 \\
    \midrule[0.5pt]
From Scratch & \multicolumn{5}{c}{90.32} \\
    \bottomrule[1pt]
    \end{tabular}}
    \label{tab:different_ms}
    } 

 \end{table}

\begin{table}[!t]

\centering
    \centering
    \setlength{\tabcolsep}{1.1mm}{
    \footnotesize
    \caption{ResNet18, 50, 152 results (evaluated on ModelNet 3D Warehouse) of finetuing the mean and variance in batch normalization layers.}
    \begin{tabular}{ m{44mm}<{\centering}| m{22mm}<{\centering}  m{22mm}<{\centering}  m{22mm}<{\centering} }
    \toprule[1pt]
    \textbf{Layers} &  \textbf{ResNet18} & \textbf{ResNet50} & \textbf{ResNet152} \\
    \midrule[0.5pt]
IO & 71.03 & 81.20 & 64.63 \\
IOms & 81.89 & 82.94 & 82.66 \\
IOmsWb & 88.75 & 89.87 & 90.44 \\
    \midrule[0.5pt]
    From Scratch & 90.39 & 90.32 & 90.28 \\
    \bottomrule[1pt]
    \end{tabular}
    \label{tab:depth_ms}
    }

 \end{table}

\subsection{Ablation study of inflating towards different directions.}

\begin{figure*}[!t]
    \centering
    \includegraphics[width=.6\textwidth]{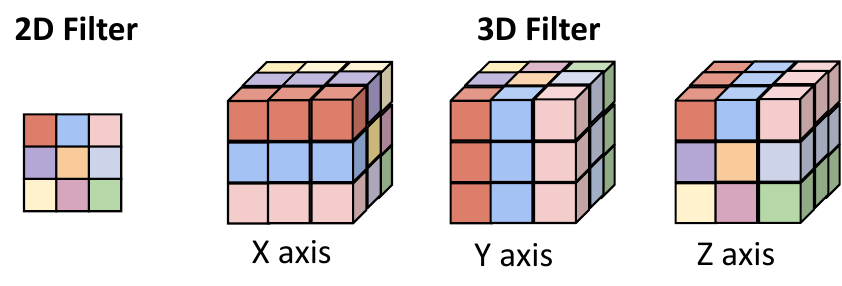}
    \caption{Visualization of filter inflation along different axis.}
    \label{xyz_inflate}
\end{figure*}

We conduct experiments of inflating filters along different directions with the illustration figure shown in Figure \ref{xyz_inflate} and the results shown in Table \ref{xyz}. We find that the performance is different when using different inflation methods. In particular, with ResNet50 pretrained on ImageNet1K, inflating along the x axis and the y axis leads to better performance compared with inflating along $z$ axis for both FIP-IO and FIP-IO+BN. More importantly, the minimally finetuned FIP-IO+BN with inflating along the $x$ and $y$ axis even surpasses the training-from-scratch.

\begin{table}[!t]
    \centering
    \setlength{\tabcolsep}{1.1mm}{
    \footnotesize
    \caption{ModelNet 3D Warehouse results (Top-1 accuracy) of partially finetuning ResNet50 pretrained on ImageNet1K with inflation along the $x, y, z$ axis.}
    \begin{tabular}{ m{30mm}<{\centering}| m{20mm}<{\centering}  m{20mm}<{\centering}  m{20mm}<{\centering} }
    \toprule[1pt]
    \textbf{Method}
    & \textbf{x axis} & \textbf{y axis} & \textbf{z axis}\\
    \midrule[0.5pt]
    FIP-IO & 82.17 & 81.73 & 81.20 \\
    FIP-IO+BN & 90.44 & 90.84 & 89.87 \\ 
    \midrule[0.5pt]
    From Scratch & \multicolumn{3}{c}{90.32} \\
    \bottomrule[1pt]
    \end{tabular}
    \label{xyz}
    }
\end{table}

\subsection{Ablation study of loading different stages of the  image-pretrained model.}
We investigate the effect of loading different subsets of stages. The results are shown in Table \ref{stages}. In detail, we load the pretrained weights partially while keeping the other weights randomly initialized. We observe that excluding the weights of the first stage achieves the best performance, bringing 0.77 points improvement.

\begin{table}[!t]
    \centering
    \footnotesize
    \caption{ModelNet 3D Warehouse results (Top-1 accuracy) of finetuning ResNet50 pretrained on ImageNet1K with only subset of stages loaded.
}
    \label{shape-texture-more}
    \begin{tabular}{ p{20mm}<{\centering}| p{12mm}<{\centering}  p{12mm}<{\centering}  p{12mm}<{\centering}  p{12mm}<{\centering} p{12mm}<{\centering}  p{12mm}<{\centering}
    p{12mm}<{\centering}}
    \toprule[1pt]
    \textbf{Loaded Stages} &  \textbf{1} & \textbf{1 2} & \textbf{1 2 3} & \textbf{1 2 3 4} & \textbf{2 3 4} & \textbf{3 4} & \textbf{4}\\
    \midrule[0.5pt]
    FIP-ALL & 89.91 & 90.36 & 90.64 & 90.92 & 91.09 & 90.19 & 89.99 \\
     \midrule[0.5pt]
    From Scratch & \multicolumn{7}{c}{90.32} \\
    \bottomrule[1pt]
    \end{tabular}
    \label{stages}
\end{table}

\subsection{Stability analysis of the semi-supervised experiment.}
For the semi-supervised experiment, we change the random seed and calculated the mean and standard deviation of three trials for each setting as shown in Table \ref{semi_error}.

\subsection{Neural Collapse in the Embedding Layer.}\label{sec:nc}

\cite{Papyan24652} characterized neural collapse as training dynamics of overparameterized neural networks in which the feature embeddings of samples from the same class tend to concentrate around their class means. In this section, we briefly define neural collapse and evaluate it in our current setting. We refer the reader for additional details in~\cite{Papyan24652,galanti2022on}.

Suppose we have a classification problem, in which we are provided with a training dataset $S = \cup^{C}_{c=1} S_c = \cup^{C}_{c=1} \{(x_{ci},c)\}^{m_0}_{i=1}$ split into classes. We would like to train a neural network $h = e \circ q$, with $q:\R^n \to \R^p$ and $e:\R^p \to \R^C$ is a linear layer. The neural network is trained by minimizing cross-entropy loss between the one-hot encodings of its labels of samples in $S$ and its logits. For additional details, see Section~\ref{sec:setup}. 

Several definitions of neural collapse have been proposed in the literature.  In this paper we work with a relatively simple definition that has been proposed in~\cite{galanti2022on}. We start by defining the \textit{class-distance normalized variance} (CDNV), which is a measure of clusterability of the feature embeddings. For a given feature map $q: \R^n \to \R^p$ and two distributions $Q_1,Q_2$ (of samples from two different classes) over $\cX$, the \footnote{The CDNV can be extended to finite sets $S_1, S_2 \subset \cX$ by defining $V_q(S_1,S_2)=V_q(U[S_1],U[S_2])$.} is defined in the following manner
\begin{equation}
V_q(Q_1,Q_2) = \frac{\Var_q(Q_1) + \Var_q(Q_2)}{2\|\mu_q(Q_1)-\mu_q(Q_2)\|^2}.
\end{equation}
Essentially, this quantity measures to what extent the deviations of the embeddings $q(x)$ of samples coming from $Q_1$ and $Q_2$ are smaller than the distance between their means. Intuitively, if the deviations are very small in comparison with the distances, then we expect the embeddings to be clustered with respect to their class labels. Note, that this quantity is also scale-invariant, i.e., if we multiply $q$ by $\alpha \neq 0$, then, the CDNV would not change for any pair $Q_1, Q_2$.  

According to the definition in~\cite{galanti2022on}, \textit{neural collapse} is defined in the following manner
\begin{equation}
\lim_{t\to \infty}\Avg_{i\neq j \in [l]}[V_{q_t}(S_i,S_j)] = 0,
\end{equation}
where $q_t$ is the embedding function after $t$ epochs of training $e\circ q$. Intuitively, during train time, the feature embeddings of samples of the same class tend to concentrate around their class-means in comparison with their distance from the other classes. 

% Certain theoretical results~\cite{Papyan24652,zhu2021geometric,mixon2020neural,Poggio2020ImplicitDR,poggio2020explicit} also identify settings in which neural collapse holds for training, or the feature map $f$ induced by the global minima of the training loss satisfies $\Avg_{i\neq j} V_f(\tS_i, \tS_j)=0$ (for large enough sample sizes). 

\paragraph{Evaluating neural collapse.} In this experiment we measure the clusterability of the feature embeddings of the penultimate layer of the pretrained model into classes, on both the source/pretraining task (i.e., ImageNet1K) and the target task (e.g., ModelNet40). In this experiment, we train a classifier of the form $\tih = \tie \circ q = \tie \circ f \circ \tig$ on ImageNet1K. We used ResNet50 as the architecture of $h$, where $\tig$ is the first convolutional layer of the model and $\tie$ is the top linear layer of the model. As a second step, we replace $\tie$ and $\tig$ with neural networks $e$ (a linear layer) and $g$ (a linear layer) and train them, along with retraining the batch normalization parameters of $f$ on ModelNet40, resulting in a function $e \circ f' \circ g$. We denote by $\tS^{tr} = \cup^{k}_{c=1} \tS^{tr}_c$ the source training dataset and by $S^{tr} = \cup^{l}_{c=1} S^{tr}_c$ the target training dataset. Here, $\tS^{tr}_c$ and $S^{tr}_c$ are the samples associated with the $c$th class. We also denote by $\tS^{val} = \cup^{k}_{c=1} \tS^{val}_c$ and $S^{val} = \cup^{k}_{c=1} S^{val}_c$ the corresponding validation datasets.

To measure the degree of clusterability of the feature embeddings, we consider multiple applications of the averaged CDNV. For measuring the clusterability of the features on the source task, we consider the CDNV of $f \circ \tig$ on the source train and validation datasets: $\Avg_{i\neq j \in [k]}[V_{f \circ \tig}(\tS^{tr}_i,\tS^{tr}_j)]$ and $\Avg_{i\neq j \in [k]}[V_{f \circ \tig}(\tS^{val}_i,\tS^{val}_j)]$. These results are reported in Table~\ref{cdnv_high}. Similarly, we also measure the CDNV of $f' \circ g$ on the source train and validation datasets: $\Avg_{i\neq j \in [k]}[V_{f' \circ g}(\tS^{tr}_i,\tS^{tr}_j)]$ and $\Avg_{i\neq j \in [k]}[V_{f' \circ g}(\tS^{val}_i,\tS^{val}_j)]$. In the main text, we report the CDNV on the validation set for $e \circ f' \circ g$ which reflects the property of the embedding's clusterability in a low-dimensional space. 

As can be seen in Table~\ref{cdnv_high}, across all of the experiments, the values of the CDNV are lower than $1$, meaning that the standard deviations of the embeddings per class are smaller in comparison with the distances between class means. Therefore, we encounter a scenario where the embeddings of samples are fairly separated into classes. In addition, we observe that the degree of collapse generalizes well to new samples, as the CDNV on the train and validation data are relatively similar. %In addition, when comparing the performance of FIP-IO+BN with ImageNet1K and ImageNet21K pretraining, we observe that with more pretraining we get improved clustering on validation set.

\begin{table}[!t]
    \centering
    \setlength{\tabcolsep}{1.1mm}{
    \footnotesize
    \caption{Stability analysis of semi-supervised distillation experiments on top of ResNet34 on the ModelNet 3D Warehouse dataset.}
    \begin{tabular}{ m{14mm}<{\centering}| m{25mm}<{\centering}  m{25mm}<{\centering}  m{25mm}<{\centering} m{25mm}<{\centering}}
    \toprule[1pt]
    \textbf{Few-shot}
    & \textbf{From scratch} & \textbf{PointInfoNCE} & \textbf{Hardest Contrastive} & \textbf{ImageNet1K pretrain (Ours)}\\
    \midrule[0.5pt]
    10-shot & 72.7$\pm$1.2 & 74.3$\pm$1.3 \textbf{(+1.6)} & 73.9$\pm$1.1 \textbf{(+1.2)} & 74.6$\pm$1.1 \textbf{(+1.9)}\\
    5-shot & 62.2$\pm$0.7 & 64.5$\pm$1.4 \textbf{(+2.3)} & 65.0$\pm$2.1 \textbf{(+2.8)} & 65.4$\pm$1.4 \textbf{(+3.2)}\\
    1-shot & 30.8$\pm$2.4 & 36.5$\pm$1.8 \textbf{(+5.7)} & 35.9$\pm$1.0 \textbf{(+5.1)} & 38.3$\pm$2.0 \textbf{(+7.5)}\\
    \bottomrule[1pt]
    \end{tabular}
    \label{semi_error}
    }
\end{table}

\begin{table}[!t]
    \centering
    \setlength{\tabcolsep}{1.1mm}{
    \footnotesize
    \caption{CDNV of the pretrained models on ImageNet1K Training/validation set, and CDNV of training from scratch and FIP-IO+BN on ModelNet 3D Warehouse training/validation set}
    \begin{tabular}{ m{25mm}<{\centering}|  m{18mm}<{\centering}|  m{20mm}<{\centering} m{26mm}<{\centering} m{26mm}<{\centering}}
    \toprule[1pt]
    \textbf{Models}
    &  \textbf{ImageNet1K} &\textbf{From scratch} & \textbf{FIP-IO+BN ImageNet1K} & \textbf{FIP-IO+BN ImageNet21K}\\
    \midrule[0.5pt]
    Training CDNV& 0.63 &0.37 & 0.71 & 0.47 \\
    Validation CDNV& 0.66 &  0.43 & 0.68 & 0.60 \\
    \bottomrule[1pt]
    \end{tabular}
    \label{cdnv_high}
    }
\end{table}

\subsection{Details of used architectures.}
\label{spp:archdetail}

\begin{lstlisting}[language=Python, caption=Pseudo code of inflated ResNet with linear input and output layers for classification.]
Class 3DRes_cls(nn.Module):
    def __init__(self, res_block):
        # res_block means the residual block as same as the conventional ResNet.
        super().__init__()
        
        self.input_layer = nn.Sequential(
            sparse_conv3d(input_dim, layer1_Idim, k=3, s=1),
            sparse_bn(layer1_Idim))
        
        self.layer1 = inflated_resnet_layer1(
            res_block, layer1_Idim, layer1_Odim)
        self.layer2 = inflated_resnet_layer2(
            res_block, layer2_Idim, layer2_Odim)
        self.layer3 = inflated_resnet_layer3(
            res_block, layer3_Idim, layer3_Odim)
        self.layer4 = inflated_resnet_layer4(
            res_block, layer4_Idim, layer4_Odim)
        
        self.output_layer = nn.Sequential(
            global_average_pooling,
            nn.Linear(layer4_Odim, class_num),
            nn.bn(class_num))
    
    def forward(self, x):
        x = self.input_layer(x)
        x = self.layer1(x)
        x = self.layer2(x)
        x = self.layer3(x)
        x = self.layer4(x)
        return self.output_layer(x)
\end{lstlisting}

\begin{lstlisting}[language=Python, caption=Pseudo code of inflated ResNet for segmentation., label=main_segarch]
Class 3DRes_seg(nn.Module):
    def __init__(self, res_block):
        # res_block means the residual block as same as the conventional ResNet.
        super().__init__()
        
        self.input_layer = nn.Sequential(
            sparse_conv3d(
                input_dim, layer1_Idim, k=3, s=1),
            sparse_bn(layer1_Idim),
            sparse_ReLU(True),
            sparse_conv3d(
                layer1_Idim, layer1_Idim, k=3, s=1),
            sparse_bn(layer1_Idim),
            sparse_ReLU(True),
            sparse_conv3d(
                layer1_Idim, layer1_Idim, k=3, s=2),
            sparse_bn(layer1_Idim),
            sparse_ReLU(True))
        
        self.layer1 = inflated_resnet_layer1(
            res_block, layer1_Idim, layer1_Odim)
        self.layer2 = inflated_resnet_layer2(
            res_block, layer2_Idim, layer2_Odim)
        self.layer3 = inflated_resnet_layer3(
            res_block, layer3_Idim, layer3_Odim)
        self.layer4 = inflated_resnet_layer4(
            res_block, layer4_Idim, layer4_Odim)
        
        self.up1 = sparse_deconv(
            layer4_Odim, layer4_Odim, k=2, s=2)
        self.decode1 = self.Sequential(
            res_block(layer4_Odim+layer3_Odim, layer3_Odim),
            res_block(layer3_Odim, layer3_Odim))
        
        self.up2 = sparse_deconv(
            layer3_Odim, layer3_Odim, k=2, s=2)
        self.decode2 = self.Sequential(
            res_block(layer3_Odim+layer2_Odim, layer2_Odim),
            res_block(layer2_Odim, layer2_Odim))

        self.up3 = sparse_deconv(
            layer2_Odim, layer2_Odim, k=2, s=2)
        self.decode3 = self.Sequential(
            res_block(layer2_Odim+layer1_Odim, layer1_Odim),
            res_block(layer1_Odim, layer1_Odim))
        
        self.up4 = sparse_deconv(
            layer1_Odim, layer1_Odim, k=2, s=2)
        self.decode4 = self.Sequential(
            res_block(layer1_Odim+layer1_Odim, layer1_Odim),
            res_block(layer1_Odim, layer1_Odim))
            
        self.output_layer = nn.Sequential(
            nn.Linear(layer1_Odim, class_num))
    
    def forward(self, x):
        x_i = self.input_layer(x)
        x1 = self.layer1(x_i)
        x2 = self.layer2(x1)
        x3 = self.layer3(x2)
        x4 = self.layer4(x3)
        
        x3_ = self.decoder1(cat(x3, self.up1(x4)))
        x2_ = self.decoder2(cat(x2, self.up2(x3_)))
        x1_ = self.decoder3(cat(x1, self.up3(x2_)))
        xi_ = self.decoder4(cat(x_i, self.up4(x1_)))
        return self.output_layer(xi_)
\end{lstlisting}

\section{Theoretical Analysis}\label{spp:theory}

To motivate our approach, in this section we provide some theoretical support for the transfer of image domain to point-cloud domain described in the main text. Instead of specifying image and point-cloud in the analysis, we begin by introducing a formal framework for analyzing transfer learning between different modalities and classes. Note that the modalities should have grounded ``relationships'', or as illustrated below, have meaningful task similarity. For example, image and point-cloud are both visual representations of the real world that share mutual characteristics, such as content and shape that could be captured using neural networks encoders.

We begin by introducing a formal framework for analyzing transfer learning between different modalities and classes. Then, we analyze a simple toy example, in which it is possible to perfectly translate one modality to the other using a linear mapping. Finally, we consider a more realistic case, in which we assume that the two domains share a `mutual semantic space' that encodes the content within samples from the two domains.

%%%%%%%%%%%%%%%%%%%%%%%%%%%%%%%%%%%%%%%%%%%%%%%%%%%%%%%%%%%%%%%
\subsection{Problem Setup}\label{sec:setup}
%%%%%%%%%%%%%%%%%%%%%%%%%%%%%%%%%%%%%%%%%%%%%%%%%%%%%%%%%%%%%%%

We extend the transfer learning setting in~\cite{galanti2022on}. We consider the problem of training a generic feature representation on a source classification task and transferring it to a target task. The two classification problems correspond to different modalities (\textit{i.e.}, two different kinds of data representations) and consist of different sets of classes. 

To model this problem, we assume that the target task is a $k$-class classification problem and the source task where the feature representation is learned on an $l$-class classification problem. Formally, the target task $T=(P,y)$ is defined by a distribution $P$ over samples $x\in \cX$, where $\cX \subset \R^{d_1}$ is the instance space, along with a function $y: \cX\to\cY_k$, where $\cY_k$ is a label space with cardinality $k$. To simplify the presentation, we use one-hot encoding for the label space, that is, the labels are represented by the unit vectors in $\R^k$, and $\cY_k=\{e_c: c=1,\ldots,k\}$ where $e_c \in \R^k$ is the $c$th standard unit vector in $\R^k$; with a slight abuse of notation, sometimes we will also write $y(x)=c$ instead of $y(x)=e_c$. For a sample $x$ with distribution $P$, we denote by $P_c(\boldsymbol{\cdot})=\P[x \in \boldsymbol{\cdot} \mid y(x)=c]$ the class distribution of $x$ given $y(x)=c$. We consider balanced classes, \textit{i.e.}, $\P[y(x)=c] = 1/k$.

\paragraph{The target task.} A classifier $h \in \mathcal{H}$ is a mapping $h:\cX\to \R_k$ that assigns a \textit{soft} label to an input point $x \in \cX$, and its performance on the target task is measured by the risk
\begin{equation}
\label{eq:loss}
L_P(h,y)=\E_{x\sim P}[\ell(h(x),y(x))],
\end{equation}
where $\ell:\R^k \times \cY_k \to [0,\infty)$ is a loss function, defined as follows $\ell(u,v) = \bI[\argmax(u) = v]$.

Our goal is to learn a classifier $h$ from some training data $S=\cup^{k}_{c=1}S_c = \cup^{k}_{c=1}\{(x_{ci},c)\}_{i=1}^n$ of $n$ independent and identically distributed (i.i.d.) samples drawn from each class $P_c$ of $P$. However, when encountering a complicated classification problem and $n$ is small, this is likely to be a hard task. To facilitate finding a good solution, we aim to find a classifier of the form $h=e \circ f \circ g$, where $f:\R^{p_1} \to \R^{p_2}$ is a feature map from a family of functions $\cF \subset \{f':\R^{p_1} \to \R^{p_2}\}$, $e \in \cE \subset \{e':\R^{p_2} \to \R^k \}$ is an affine function and $g$ is a function from a family $\cG_1 \subset \{g':\R^{d_1} \to \R^{p_1}\}$. Namely, the feature map $f$ is trained on a source problem, potentially of a different modality, where much more data is available, and then $g$ and $e$ are trained on $S$ while freezing $f$. Intuitively, $g$ and $e$ are relatively `simple' functions (\textit{e.g.}, linear layers) that are task-specific. Concretely, $e$ (the classifier) is responsible for translating $f$ into a classifier between the classes in $P$ and $g$ (the adaptor) adapts $f$ to the specific modality of $P$. 

\paragraph{The source task.} We assume that the source task helping to find $f$ is an $l$-class classification problem over a different sample space $\cX' \subset \R^{d_2}$. For example, $\cX$ could be a set of 3D point-clouds, while $\cX'$ is a set of 2D natural images. The source task $B=(\tP,\tiy)$ is defined by a distribution $\tP$ and function $\tiy:\cX' \to \cY_l$, and here we are interested in finding a classifier $\tih:\cX' \to \R^l$ of the form $\tih=\tie \circ f \circ \tig$, where $\tie \in \tcE \subset \{e':\R^p \to \R^l\}$ is an affine function over the feature space $f(\tig(\cX))=\{f(\tig(x)):x \in \cX\}$ and $\tig \in \cG_2 \subset \{g':\R^{d_2} \to \R^{p_1}\}$ is an adaptor.
Given a training dataset $\tS=\cup^{l}_{c=1} \tS_c = \cup^{l}_{c=1}\{(\tx_{ci},c)\}_{i=1}^m$, all components of the classifier, denoted by $\tig$, $f$ and $\tie$ are trained on $\tS$, with the goal of minimizing the cross-entropy loss in the source task. Following~\cite{galanti2022on} we assume that $\tS_c$ are drawn i.i.d.\ from $\tP_c$ and that $\P[\tiy(x)=c] = 1/l$. 

% In a typical setting $\tih$ is a deep neural network, $f$ is the representation in the last internal layer of the network (\textit{i.e.}, the penultimate \emph{embedding} layer), and $\tie$, the last layer of the network, is a linear map; similarly $e$ in the target problem is often taken to be linear. 

%The learned feature map $f$ is called a \emph{foundation model}~\cite{DBLP:journals/corr/abs-2108-07258} when it can be effectively used in a wide range of tasks. 

\paragraph{Tasks similarity.} In general, transferring between the source and target tasks is meaningless if the two tasks are extremely unrelated to each other. For instance, we should not expect to have any guarantee to transfer knowledge from very different tasks, such as voice separation and image segmentation. 

Intuitively, we think of the classes of the target (source) task as being of \textit{`similar character'}. To formalize this intuition, we simply assume that the target (source) classes are i.i.d.\ samples of a distribution $\mathcal{D}_1$ over $\cC_1$ ($\mathcal{D}_2$ over $\cC_2$). In~\cite{galanti2022on} they assumed that the source and target classes differ, but share the same underlying distribution, \textit{i.e.}, $\cD_1=\cD_2$. Since in this work we intend to study transfer between different modalities, $\cD_1$ need not be the same as $\cD_2$. To formally define notions of similarity between domains, we first assume the existence of an invertible mapping $F: \cC_1 \to \cC_2$, such that, $\hP_c := F(P_c) \sim \cD_2$ for $P_c \sim \cD_1$. In a sense, $\cD_1$ and $\cD_2$ share the same set of categories, encoded with different kinds of modalities. The mapping $F$ takes a certain class $P_c \in \cC_1$ and maps it to its analogous class $\hP_c \in \cC_2$ in the second domain. For a given target task $T=(P,y)$ with distribution $P$, classes $\{P_c\}^{k}_{c=1}$ and function $y:\cX \to \R^k$, we denote $A=(\hP,\hiy)$ the corresponding analogous task within the second domain, where $\hP_c = F(P_c)$ (not to be confused with the source task $B = (\tP, \tiy)$). In general, $F$ could be an arbitrary mapping between the classes. Therefore, to concretely relate between $\cD_1$ and $\cD_2$ we will have to make additional assumptions about the relationship between $P$ and $\hP$. The specific relationship between $P_c$ and $\hP_c$ will be explicitly defined near the presentation of each result. %\tg{MAKE MORE GENERIC} Instead, in order to capture the relation between $\cD_1$ and $\cD_2$, we simply assume that $g^* \circ \cD_1 = \cD_2$ for some $g^* \in \cG$. Namely, $g^* \circ P \sim \cD_2$ for $P \sim \cD_1$ (where $g \circ P$ is the distribution of $g(x)$ for $x\sim P$). Intuitively, each the classes from $\cD_2$ can be represented as a projection of $g^*$ over the classes in $\cD_1$. While $g^*$ is unknown to the learning algorithm, as we will see in Theorem~\ref{thm:transfer1} this assumption makes it possible to effectively transfer between the two domains.

\paragraph{Evaluation process.} As a next step, we would like to evaluate the performance of the pretrained feature map $f$ on the target task. To do so, we evaluate its expected performance over the distribution of binary classification target tasks 
\begin{equation}\label{eq:error}
L^k_{\cD_1}(f)=\E_{P_1\neq \dots\neq P_k \sim \cD_1}\E_{S_1,\dots,S_k} [L_P(e_S \circ f \circ g_S, y)],
\end{equation}
where $S_c\sim P^{n}_c$ and $e_S, g_S$ are the outputs of a learning algorithm that trains $e \in \cE$ and $g \in \cG_1$ to fit $e \circ f \circ g$ to the dataset $S$ while freezing $f$. For simplicity, in this work we focus on $k=2$ and denote $L_{\cD_1}(f) = L^2_{\cD_1}(f)$, even though the analysis could be readily extended to $k > 2$. Note that several implementations of mappings $S \mapsto e_S,g_S$ are possible. For simplicity, in this work, we choose $\argmax \circ e_S$ to be the \textit{`nearest empirical mean classifier'} and $g_S$ to be an empirical risk minimizer. Formally, for a given embedding function $h$, we consider the linear function $e^{S}_{h}(z) = (\langle z, 2\mu_h(S_c)\rangle - \|\mu_h(S_c)\|^2)_{c=1,2}$. In this case, $\argmax \circ e^S_h(z) = \min_{c=1,2} \|z - \mu_{h}(S_c)\|$ forms a nearest empirical mean classification rule that classifies a vector $z$ as $1$ if it is closer to the empirical embeddings mean $\mu_h(S_1) = \Avg_{x \sim S_1}[h(x)]$ than to $\mu_h(S_2) = \Avg_{x \sim S_2}[h(x)]$. In addition, $g_S = \argmin_{g \in \cG} L_X(e^S_{f\circ g} \circ f \circ g, y)$, where $X = \cup^{k}_{c=1}\{x_{ci}\}^{n}_{i=1}$ and $e_S = e_{f\circ g_S}$ is the corresponding nearest empirical mean classifier.  Notice that while the feature map $f$ is evaluated on the distribution of target tasks determined by classes taken from $\cD_1$, the training of $f$, as described above, is fully agnostic of this target. 

To summarize, in the proposed setting we train a feature map $f$ as part of a classification model $\tie \circ f$ to fit some source data corresponding to a set of source classes $\tP_1,\dots,\tP_l$ using dataset $\tS = \cup^{l}_{c=1} \tS_c$. At the second stage, $f$'s performance is evaluated against a randomly selected set of target classes $P_1,\dots,P_k$ the differ by \textit{content} (i.e., different categories) and \textit{modality} (e.g., 2D and 3D point clouds) by training an adaptor $g$ and a linear classifier $e$ based on the available data $S = \cup^{k}_{c=1} S_c$. Therefore, in this work we deal with \textit{ two modes of transfer}. First, we would like to train $f$ to be a generic feature map that can be used to distinguish between many different categories. The second deals with the ability to adapt $f$ from one modality to another using minimal efforts. Intuitively, if the pretrained feature map $f$ enjoys both qualities, we expect $L_{\cD_1}(f)$ to be small. This is what we show in Theorem~\ref{thm:transfer1}. 

%In contrast, several transfer learning methods (such as few-shot learning algorithms) have been developed which optimize $f$ not only as a function of its training data $\tS$, but also based on some properties of $\cD$, such as the number of classes and the number of samples per class in $S$.
%Perhaps surprisingly, recent studies~\citep{DBLP:conf/eccv/TianWKTI20,Dhillon2020A} demonstrate that the target-agnostic training of foundation models (or some slight variation of them, such as transductive learning) are competitve with such special purpose algorithm. In the rest of the paper we analyze this phenomenon, and provide an explanation through the recent concept of neural collapse.

\paragraph{Notation.}
Throughout the analysis, we use the following notations. For an integer $k\geq 1$, $[k]=\{1,\ldots,k\}$. For any real vector $z$, $\|z\|$ denotes its Euclidean norm. For a given set $A = \{a_1,\dots,a_n\} \subset B$ and a function $u \in \cU \subset \{u':B \to \mathbb{R}\}$, we define $u(A) = \{u(a_1),\dots,u(a_n)\}$ and $\cU(A) = \{u(A) : u \in \cU\}$. Let $Q$ be a distribution over $\cX \subset \R^d$ and $u: \cX \to \R^p$. We denote by $\mu_u(Q) = \E_{x \sim Q}[u(x)]$ and by $\Var_u(Q) = \E_{x \sim Q}[\|u(x)-\mu_u(Q)\|^2]$ the mean and variance of $u(x)$ for $x\sim Q$. For $A$ above, we denote by $\Avg^{n}_{i=1}[a_i] = \Avg A = \frac{1}{n} \sum^{n}_{i=1}a_i$ the average of $A$. For a finite set $A$, we denote by $U[A]$ the uniform distribution over $A$. We denote by $\bI:\{\textnormal{True},\textnormal{False}\} \to \{0,1\}$ the indicator function. For a given distribution $P$ over $\cX$ and a measurable function $f:\cX\to \cX'$, we denote the distribution of $f(x)$ by $f\circ P$. For two classes of functions $\cG = \{g':\R^{d_1}\to \R^{d_2}\}$ and $\cF = \{f':\R^{d_2}\to \R^{d_3}\}$, we denote $\cF\circ \cG = \{f \circ g \mid f\in \cF, g\in \cG\}$. %For a given set $S_c = \{(x_{ci},c)\}^{n}_{i=1}$ of labeled samples, we denote by $X_c =\{x_{ci}\}^{n}_{i=1}$ its unlabeled version. 

\section{Theoretical Results}

\subsection{Case 1}\label{sec:case1}

As a toy example, we first consider the case where the first domain (target) can be translated into the second domain (source), using some simple function $g^*$. For this purpose, we assume that $\cG_1$ is decomposed into $\cG_1 = \cG''_1 \circ \cG'_1$, where $\cG'_1 \subset \{g':\R^{d_1} \to \R^{d_2}\}$ and $\cG''_1 \subset \{g':\R^{d_2} \to \R^{p_1}\}$. Intuitively, the functions $g \in \cG_1$ are decomposed into sub-architectures $g'_1:\R^{d_1} \to \R^{d_2}$ and $g''_1:\R^{d_2} \to \R^{p_1}$. In addition, we assume there exists a function $g^* \in \cG'_1$, such that $g^* \circ P \sim \cD_2$ for $P \sim \cD_1$ (where $g \circ P$ is the distribution of $g(x)$ for $x\sim P$). In addition, we assume that $\cG_2 \subset \cG''_1$. Hence, for each candidate function $\tig \in \cG_2$ that could have been selected when training the classifier $h = \tie \circ f \circ \tig$, there exists a function $g = \tig \circ g^* \in \cG_1$ that maps $P_c$ into $g \circ P_c = \tig \circ \tP_c$. Intuitively, any representation $\tig \circ \tP$ of the distribution $\tP$ could be implemented as $g \circ P$ for some $g \in \cG_1$. While $g^*$ is unknown to the learning algorithm, as we will see in Theorem~\ref{thm:transfer1}, this simplifying assumption makes it possible to easily transfer between the two domains.

The following theorem provides an upper bound on the expected error of a pretrained feature map $f$ in terms of the expected CDNV between pairs of classes $\tP_i$ and $\tP_j$ from $\cD_2$ and the empirical Rademacher complexity of the class $\cH_f = \{\argmax \circ e \circ f \circ g \mid (e,g) \in \cE \times \cG_1 \}$. The Rademacher complexity is a measure of generalization that quantifies the ability of a class of functions to fit noise. Formally, for a given set $X = \{x_i\}^{n}_{i=1} \subset \R^d$ and set of functions $\cH \subset \{h' \mid h':\R^{d} \to \R\}$, the empirical Rademacher complexity (see~\cite{Mohri:2012:FML:2371238}) of $\cH$ is defined as follows 
\begin{equation}
\Rad_X(\cH) = \E_{\sigma}\left[\sup_{h \in \cH} \frac{1}{n}\sum^{n}_{i=1} \sigma_i \cdot f(x_i) \right],
\end{equation}
where $\sigma = (\sigma_1,\dots,\sigma_n)$ are i.i.d. uniformly distributed over $\{\pm 1\}$. The empirical Rademacher complexity can lead to tighter bounds than those based on other measures of complexity such as the VC-dimension~\cite{koltchinskii2004rademacher}. It also has the added advantage that it is data-dependent and can be measured from finite samples.

\begin{theorem}\label{thm:transfer1}
In the setting above. For any tuple $(f,\tig) \in \cF \times \cG$, we have:
\begin{equation*}
\begin{aligned}
L_{\cD_1}(f) &~\leq~ 16\mathbb{E}_{\tP_1 \neq \tP_2 \sim \mathcal{D}_2} \left[\frac{V_{f\circ \tig}(\tP_1,\tP_2)}{s(f\circ \tig,\tP_1,\tP_2)}\right] + 6\sqrt{\frac{\log(4n)}{2n}} + \frac{2}{n} \\
&~~\quad+2\E_{P_c\sim \cD_1}\E_{X_c\sim P^{n}_c}\left[\Rad_{X_c}(\cH_f)\right],
\end{aligned}
\end{equation*}
where $s(f,P_1,P_2)=p_2$ if $f \circ P_1$ and $f \circ P_2$ are spherically symmetric and $s(f,P_1,P_2)=1$ otherwise. 
\end{theorem}

The theorem above provides an upper bound on the expected error $L_{\cD_1}(f)$ of a pretrained feature map $f$ against classification tasks generated using classes from $\cD_1$. The bound holds uniformly for any pair $(f,\tig) \in \cF \times \cG$. Hence, we think of $f$ and $\tig$ as the pretrained functions that were obtained by training the classifier $\tie \circ f \circ \tig$ to fit the source data $\tS = \cup^{l}_{c=1} \tS_c$. The bound is decomposed into several parts. The first term is (approximately) proportional to the CDNV between pairs of randomly selected classes from $\cD_2$. Namely, it measures the extent of neural collapse we encounter between randomly selected pairs of source classes $\tP_1\neq \tP_2 \sim \cD_2$. 

As mentioned in Section~\ref{sec:nc} in the regime of neural collapse, we expect $\Avg_{i\neq j \in [l]} \left[V_{f \circ \tig}(\tS_i,\tS_j)\right]$ to be small. In addition, by Propositions~1 in~\cite{galanti2022on}, under certain circumstances if the number of samples per source class $m$ is large enough, we also expect $\Avg_{i\neq j \in [l]} \left[V_{f \circ \tig}(\tP_i,\tP_j)\right]$ to be small. Finally, by Propositions~2 in~\cite{galanti2022on}, if the number of source classes $l$ is also large, we should also expect $\mathbb{E}_{\tP_i \neq \tP_j \sim \mathcal{D}_2}[V_{f \circ \tig}(\tP_i,\tP_j)]$ to be small. As a side note, a smaller bound is obtained when $f \circ \tig$ projects the classes $\tP_c \sim \cD_2$ to spherically symmetric distributions as $s(f \circ \tig, \tP_1,\tP_2) = p_2$ if $f \circ \tig \circ \tP_c$ are spherically symmetric. For estimations of $\Avg_{i\neq j \in [l]} \left[V_{f \circ \tig}(\tS_i,\tS_j)\right]$ and $\Avg_{i\neq j \in [l]} \left[V_{f \circ \tig}(\tP_i,\tP_j)\right]$, see Section~\ref{sec:nc}, in which $\tS_c$ is denoted by $\tS^{tr}_c$ and the class distributions $\tP_c$ are approximated with validation sets $\tS^{val}_c$.

The second group of terms includes the (expected) Rademacher complexity of the class $\cH_f$ and additional terms that scale as $\mathcal{O}\left(\sqrt{\log(n)/n}\right)$. The Rademacher complexity $\Rad_S(\cH)$ of a class $\cH$ of neural network classifiers $h:\R^d \to \{0,1\}$ typically scales as $\mathcal{O}\left(\sqrt{N/n} \right)$, where $N$ polynomially depends on the number of trainable parameters and $n$ is the number of samples. Therefore, in standard settings, we expect $\Rad_S(\cH_f)$ to scale as $\mathcal{O}\left(\sqrt{N/n} \right)$, where $N$ polynomially depends on the number of parameters in $e$ and $g$. On the other hand, a standard Rademacher complexity generalization bound would yield a dependence on the number of parameters existing in the full network $e \circ f \circ g$ (including the ones in $f$). Since typically $f$ contains most of the trainable parameters in the neural network, this allows us to significantly reduce the sample complexity of the target task.

\begin{proof}
To prove this theorem, we fix a target task $T=(P,y)$ with a pair of classes $P_1,P_2$ and a pretrained feature map $f \circ \tig$. 
By (cf.~\cite{Mohri:2012:FML:2371238}, Theorem~3.5), for any $c=1,2$, with probability at least $1-\frac{1}{4n}$ over the selection of $S_c$, for any pair $(e,g) \in \cE\times \cG_1$, we have 
\begin{equation}
\begin{aligned}
\vert L_{P_c}(e \circ f \circ g, y) - L_{X_c}(e\circ f \circ g, y) \vert \leq 2\Rad_{X_c}(\cH_f) + 3\sqrt{\frac{\log(4n)}{2n}},
\end{aligned}
\end{equation}
where $X_c$ consists of the samples in $S_c$ excluding their labels and $X = X_1\cup X_2$. By union bound over $c=1,2$, with probability at least $1-\frac{1}{2n}$ over the selection of $S$, the following inequality holds uniformly for all $(e,g) \in \cE \times \cG_1$,
\begin{equation}
\begin{aligned}
\vert L_{P}(e \circ f \circ g, y) - L_{X}(e \circ f \circ g, y) \vert ~\leq~ \sum_{c=1,2}\Rad_{X_c}(\cH_f) + 3\sqrt{\frac{\log(4n)}{2n}}.
\end{aligned}
\end{equation}
Hence, with probability at least $1-\frac{1}{2n}$ over the selection of $S$,
\begin{equation}
\begin{aligned}
L_P(e_S \circ f \circ g_S, y) &~\leq~ L_X(e_S \circ f \circ g_S, y)\\ &~~\quad+ \sum_{c=1,2}\Rad_{X_c}(\cH_f) + 3\sqrt{\frac{\log(4n)}{2n}}. \\
\end{aligned}
\end{equation}
Let $\cE' = \left\{e(z) = (-\|z - \mu_c\|^2)_{c=1,2} = (\langle z, 2\mu_c\rangle - \|\mu_c\|^2)_{c=1,2} \mid \mu_1,\mu_2 \in \R^{p_2}\right\} \subset \cE$. In addition, we let $e_{P}(z) = (-\|z - \mu_{f \circ \tig \circ g^*}(P_c)\|^2)_{c=1,2}$. Since $\tig \circ g^* \in \cG_2 \circ \cG'_1 \in \cG''_1\circ \cG'_1 = \cG_1$, we have
\begin{equation}
\begin{aligned}
L_X(e_S \circ f \circ g_S, y) &~=~ \inf_{g \in \cG_1} \inf_{e \in \cE'} L_X(e \circ f \circ g, y) \\
&~\leq~ \inf_{e \in \cE'} L_X(e \circ f \circ \tig \circ g^*, y) \\
&~\leq~ L_X(e_{P} \circ f \circ \tig \circ g^*, y).
\end{aligned}
\end{equation}
In particular, since the loss function is bounded in $[0,1]$ and the above event holds with probability at least $1-\frac{1}{2n}$, by taking expectation over $S$, we obtain the following inequality
\begin{equation}
\begin{aligned}
&\mathbb{E}_{S}\left[L_P(e_S \circ f \circ g_S, y)\right] \\
&~\leq~ \mathbb{E}_{S}\left[L_S(e_P \circ f \circ \tig \circ g^*, y)\right] + 3\sqrt{\frac{\log(4n)}{2n}} + \frac{1}{2n} + \sum_{c=1,2}\Rad_{X_c}(\cH_f) \\
&~=~ \mathbb{E}_{S}\left[L_S(e_P \circ f \circ \tig \circ g^*, y)\right] + 3\sqrt{\frac{\log(4n)}{2n}} + \frac{1}{2n} + \sum_{c=1,2}\Rad_{X_c}(\cH_f).
\end{aligned}
\end{equation}
Finally, we can take expectation over the selection of $P_1,P_2$ on both sides of the inequality to obtain that
\begin{equation}
\begin{aligned}
L_{\cD_1}(f) &~=~ \mathbb{E}_{P_1\neq P_2 \sim \cD_1}\mathbb{E}_{S}\left[L_P(e_S \circ f \circ g_S, y)\right] \\
&~\leq~ \mathbb{E}_{P_1\neq P_2}\left[L_P(e_P \circ f \circ \tig \circ g^*,y)\right] \\ 
&~~\quad + 2\mathbb{E}_{P_1\neq P_2\sim \cD_1}\E_{X_c}\left[\Rad_{X_c}(\cH_f)\right] + 3\sqrt{\frac{\log(4n)}{2n}} + \frac{1}{2n}.
\end{aligned}
\end{equation}
Finally, since for any given distribution $P$, we have $g^* \circ P \sim \mathcal{D}_1$ for $P \sim \mathcal{D}_2$, by Proposition~5 in~\cite{galanti2022on}, we obtain that 
\begin{equation}
\begin{aligned}
\mathbb{E}_{P_1\neq P_2}\left[L_P(e_P \circ f \circ \tig \circ g^*,y)\right] 
&~=~ \mathbb{E}_{\hP_1\neq \hP_2}\left[L_{\hP}(e_P \circ f \circ \tig, \hiy)\right] \\
&~\leq~ \frac{16 V_{f\circ \tig}(\hP_1,\hP_2) }{s(f\circ \tig,\hP_1,\hP_2)}.
\end{aligned}
\end{equation}
\end{proof}

\subsection{Case 2}\label{sec:case2}

In the previous section, we assumed existence of a mapping $g^* \in \cG'_1$, such that $g^* \circ P_c \sim \cD_2$ for $P_c \sim \cD_1$. However, this assumption is typically violated in practice~\cite{pmlr-v48-reed16,pmlr-v37-xuc15,DBLP:journals/corr/abs-1907-01341}. Therefore, following the Unsupervised Domain Adaptation literature~\cite{bendavid,DBLP:conf/alt/Mansour09}, we use a relaxed assumption that there is a `shared representation space' for both domains. Informally, the two domains can be mapped to a shared space, in which classification into classes is possible. Variations of this assumption are algorithmically and theoretically used in multiple areas of computer vision~\cite{huang2018munit,Benaim2019DomainIntersectionDifference,CycleGAN2017,radford2021learning,Liu2021ContrastiveMF,ALBEF}.

Formally, we assume the existence of two mappings $g^*\in \cG_1$ and $\tig^* \in \cG_2$ that satisfy $g^* \circ P_c \approx \tig^* \circ \hP_c$ in expectation over $P_c \sim \cD_1$. To formalize this assumption, we make use of the notion of discrepancy~\cite{bendavid,Chazelle2000TheDM}. Namely, for a given set $\cV$ of functions $v:\cX \to \R$, we define the discrepancy between two distributions $P_1$ and $P_2$ over $\cX$, as $\textnormal{disc}_{\cV}(P_1,P_2) = \sup_{h \in \cV} \vert \E_{x\sim P_1}[h(x)] - \E_{x\sim P_2}[h(x)] \vert$. The discrepancy, or Integral Probability Metric (IPM)~\cite{muller}, is a pseudo-metric between distributions. Namely, we (implicitly) assume that $\E_{P_c \sim \cD_1}\left[\textnormal{disc}_{\cV}(g^* \circ P_c, \tig^* \circ \hP_c)\right]$ is small, where $\cV$ is some class of binary functions (to be defined in the proof). %A special case of IPMs is the 1-Wasserstein distance, where $\cV$ consists of all $1$-Lipschitz functions.

To capture the intuition that one can classify the samples into classes from the representation space $g^*(\cX) \approx \tig^*(\cX')$, we assume that the following term
\begin{equation}
\E_{P_1\neq P_2} \left[\inf_{(e, f) \in \cE\times \cF}  L_P(e \circ f \circ g^*, y) + L_{\hP}(e \circ f \circ \tig^*, \hiy) \right]
\end{equation}
is small. In words, in expectation over the selection of $P$ and $\hP$, the correct labels $y(x)$ and $\hiy(x)$ can be simultaneously recovered using classifiers $e \circ f \circ g^*$ and  $e \circ f \circ \tig^*$, for some $e \circ f \in \cE \circ \cF$. 

Finally, as a technicality, we assume that the pretrained adaptor $\tig$ can be represented as $\tig = u \circ \tig^*$, for some function $u \in \cU$, where $\cU \subset \{u':\R^{p_1}\to \R^{p_1}\}$, such that, $\cU \circ \cG_1 \subset \cG_1$. For example, if $\cG_1$ is a class of neural networks, ending with a linear layer and $\cU$ is the set of linear mappings $u:\R^{d_1}\to \R^{d_1}$, then indeed we have $\cU \circ \cG_1 \subset \cG_1$.

\begin{theorem}\label{thm:transfer2}
In the setting above. For any pair $(f,\tig) \in \cF  \times \cG$, such that $\tig = u \circ \tig^*$, we have:
\begin{equation*}
\begin{aligned}
L_{\cD_1}(f) &\leq 16\E_{\tP_1 \neq \tP_2 \sim \mathcal{D}_2}\left[\frac{V_{f \circ \tig}(\tP_i,\tP_j)}{s(f \circ \tig, \tP_1, \tP_2)}\right] + 2\E_{P_c}\E_{X_c \sim P^{n}_c}[\Rad_{X_c}(\cH_f)] \\
&\quad+ \E_{P_c}[\textnormal{disc}_{\cV}(g^* \circ P_c, \tig^* \circ \hP_c)] + 3\sqrt{\frac{\log(4n)}{2n}} + \frac{1}{2m} \\
&\quad+ \E_{P_1\neq P_2} \left[\inf_{(e, f) \in \cE\times \cF}  L_P(e \circ f \circ g^*, y) + L_{\hP}(e \circ f \circ \tig^*, \hiy) \right],
\end{aligned}
\end{equation*}
where $s(f,P_1,P_2)=p_2$ if $f \circ P_1$ and $f \circ P_2$ are spherically symmetric and $s(f,P_1,P_2)=1$ otherwise.
\end{theorem}

The above theorem provides an upper bound on the expected error of the pretrained feature map $f$ against binary classification target tasks $T=(P,y)$. In this theorem, we assume that $f$ has been trained along with an adaptor $\tig$ that can be represented as $u \circ \tig^*$, where $u$ is a linear mapping. 

The upper bound is decomposed into multiple parts. Similar to the previous bound it sums the expected CDNV between pairs of classes $\tP_1, \tP_2$, the Rademacher complexity of $\cH_f$ and additional terms scaling as $\mathcal{O}(\sqrt{\log(n)/n})$. As discussed in Section~\ref{sec:case1}, we expect these terms to be small. In addition, the bound also includes the expected discrepancy between $g^* \circ P_c$ and $\tig^* \circ \hP_c$ that measures to what extent the two adaptors $g^*$ and $\tig^*$ can map the distributions $P_c$ and $\hP_c$ to the same space. Finally, the last term measures to what extent we can actually recover the class label from representations taken from the shared space $g^* \circ P_c\approx \tig^* \circ \hP_c$. As mentioned above, we explicitly assume that these terms are small. Intuitively, these terms are small when the two domains share a mutual semantic space $g^* \circ P_c\approx \tig^* \circ \hP_c$ that encodes the content within samples from the two domains. 

The proof of this theorem is based on the analysis of~\cite{galanti2022on}, the theory of Unsupervised Domain Adaptation~\cite{bendavid,DBLP:conf/alt/Mansour09,DBLP:conf/colt/MansourMR09} and Rademacher complexities~\cite{Mohri:2012:FML:2371238}.

\begin{proof}
Let $(\hat{e},\hat{f}) \in \cE \times \cF$ be any pair of functions. Let $u$ be a linear mapping, such that, $\tig = u \circ \tig^*$. To prove this theorem, we start by considering a pair of target classes $P_1,P_2$. Since the loss function is bounded in $[0,1]$, for any $c=1,2$, by (cf.~\cite{Mohri:2012:FML:2371238}, Theorem~3.3) with probability at least $1-\frac{1}{4m}$ over the selection of $S_c$, for any pair $(e,g) \in \cE\times \cG$, we have 
\begin{equation}
\begin{aligned}
L_{P_c}(e \circ f \circ g, y)  
&~\leq~ L_{X_c}(e\circ f \circ g, y) + 3\sqrt{\frac{\log(4m)}{2m}} + 2\Rad_{X_c}(\cH_f).
\end{aligned}
\end{equation}
By union bound over $c=1,2$, with probability at least $1-\frac{1}{2m}$ over the selection of $S = S_1\cup S_2$, for any pair $(e,g) \in \cE\times \cG$, we have 
\begin{equation}
\begin{aligned}
L_{P}(e \circ f \circ g, y) &~\leq~ L_{S}(e\circ f \circ g, y) + 3\sqrt{\frac{\log(4m)}{2m}} + \sum_{c=1,2}\Rad_{S_c}(\cH_f).
\end{aligned}
\end{equation}
Hence, with probability at least $1-\frac{1}{2m}$ over the selection of $S$,
\begin{equation}
\begin{aligned}
L_P(e_S \circ f \circ g_S, y) &~\leq~ L_S(e_S \circ f \circ g_S, y) + 3\sqrt{\frac{\log(4m)}{2m}} \\
&~~\quad+ \sum_{c=1,2}\Rad_{S_c}(\cH_f).
\end{aligned}
\end{equation}
Let $\cE' = \left\{e(z) = (-\|z - \mu_c\|^2)_{c=1,2} = (\langle z, 2\mu_c\rangle - \|\mu_c\|^2)_{c=1,2} \mid \mu_1,\mu_2 \in \R^{p_2}\right\} \subset \cE$. In addition, we let $e_{\hP}(z) = (-\|z - \mu_{f \circ \tig}(\hP_c)\|^2)_{c=1,2}$. We notice that 
\begin{equation}
\begin{aligned}
L_S(e_S \circ f \circ g_S, y) &~=~ \inf_{g \in \cG_1} \inf_{e \in \cE'} L_S(e \circ f \circ g, y) \\
&~\leq~ \inf_{e \in \cE'} L_S(e \circ f \circ u \circ g^*, y) \\
&~\leq~ L_S(e_{\hP} \circ f \circ u \circ g^*, y).
\end{aligned}
\end{equation}
Since the loss function is bounded in $[0,1]$ and the above event holds with probability at least $1-\frac{1}{2m}$, by taking expectation over $S$, we have the following
\begin{equation}
\begin{aligned}
\E_{S}\left[L_P(e_S \circ f \circ g_S, y)\right] 
&\leq L_P(e_{\hP} \circ f \circ u \circ g^*, y) + 3\sqrt{\frac{\log(4m)}{2m}} + \frac{1}{2m} \\
&\quad+ \sum_{c=1,2}\E_{S_c}\left[\Rad_{S_c}(\cH_f)\right] \\
\end{aligned}
\end{equation}
In addition,
\begin{equation}
\begin{aligned}
L_P(e_{\hP} \circ f \circ u \circ g^*, y) &~\leq~ L_P(e_{\hP} \circ f \circ u \circ g^*, \hat{e} \circ \hat{f} \circ g^*) \\
&~~\quad + L_P(\hat{e} \circ \hat{f} \circ g^*, y) \\
&~\leq~  L_{\hP}(e_{\hP} \circ f \circ u \circ \tig^*, \hat{e} \circ \hat{f} \circ \tig^*) \\
&~~\quad + L_P(\hat{e} \circ \hat{f} \circ g^*, y) \\
&~~\quad+ \textnormal{disc}_{\cV}(g^* \circ P, \tig^* \circ \hP) \\
&\leq  L_{\hP}(e_{\hP} \circ f \circ u \circ \tig^*, \hiy) \\
&~~\quad+ L_{\hP}(\hat{e} \circ \hat{f} \circ \tig^*, \hiy) + L_P(\hat{e} \circ \hat{f} \circ g^*, y) \\
&~~\quad+ \textnormal{disc}_{\cV}(g^* \circ P, \tig^* \circ \hP),
\end{aligned}
\end{equation}
where $\cV = \{\bI[e_1 \circ f_1 \circ u_1 \neq e_2 \circ f_2] \mid e_1,e_2 \in \cE, f_1,f_2 \in \cF, u_1 \in \cU\}$. In particular, we can take infimum over $(\hat{e},\hat{f}) \in \cE\times \cF$ to obtain
\begin{equation}
\begin{aligned}
L_P(e_{\hP} \circ f \circ u \circ g^*, y) &~\leq~  L_{\hP}(e_{\hP} \circ f \circ u \circ \tig^*, \hiy) \\
&~~\quad+ \inf_{e,f} \left[L_{\hP}(e \circ f \circ \tig^*, \hiy) + L_P(e \circ f \circ g^*, y)\right] \\
&~~\quad+ \textnormal{disc}_{\cV}(g^* \circ P, \tig^* \circ \hP) \\
&~=~  L_{\hP}(e_{\hP} \circ f \circ \tig, \hiy) \\
&~~\quad+ \inf_{e,f} \left[L_{\hP}(e \circ f \circ \tig^*, \hiy) + L_P(e \circ f \circ g^*, y)\right] \\
&~~\quad+ \textnormal{disc}_{\cV}(g^* \circ P, \tig^* \circ \hP).
\end{aligned} 
\end{equation}
Next, we can take expectation over the selection of $P$ on both sides of the inequality to obtain that
\begin{equation}
\begin{aligned}
&\E_{P_1\neq P_2} \left[L_P(e_{\hP} \circ f \circ u \circ g^*, y)\right] \\
&~\leq~ \E_{P_1\neq P_2} \left[L_{\hP}(e_{\hP} \circ f \circ \tig, \hiy)\right] \\
&~~\quad+ \E_{P_1\neq P_2} \left[\inf_{e,f} \left[L_{\hP}(e \circ f \circ \tig^*, \hiy) + L_P(e \circ f \circ g^*, y)\right]\right] \\
&~~\quad+ \E_{P_1\neq P_2} \left[\textnormal{disc}_{\cV}(g^* \circ P, \tig^* \circ \hP)\right] \\
&~=~  \E_{\hP_1\neq \hP_2} \left[L_{\hP}(e_{\hP} \circ f \circ \tig, \hiy)\right] \\
&~~\quad+ \E_{P_1\neq P_2} \left[\inf_{e,f} \left[L_{\hP}(e \circ f \circ \tig^*, \hiy) + L_P(e \circ f \circ g^*, y)\right]\right] \\
&~~\quad+ \E_{P_1\neq P_2} \left[\textnormal{disc}_{\cV}(g^* \circ P, \tig^* \circ \hP)\right]
\end{aligned}
\end{equation}
We note that
\begin{equation}
\begin{aligned}
&\textnormal{disc}_{\cV}(g^* \circ P, \tig^* \circ \hP) \\
&~=~ \sup_{v \in \cV} \left\vert \E_{z \sim g^* \circ P}[v(z)] - \E_{z \sim \tig^* \circ \hP}[v(z)] \right\vert \\
&~=~ \sup_{v \in \cV} \left\vert \frac{1}{2}\sum_{c=1,2}\E_{z \sim g^* \circ P_c}[v(z)] - \frac{1}{2}\sum_{c=1,2}\E_{z \sim \tig^* \circ \hP_c}[v(z)] \right\vert \\
&~\leq~ \frac{1}{2}\sum_{c=1,2}\sup_{v \in \cV} \left\vert \E_{z \sim g^* \circ P_c}[v(z)] - \E_{z \sim \tig^* \circ \hP_c}[v(z)] \right\vert \\
&~\leq~ \frac{1}{2}\sum_{c=1,2}\textnormal{disc}_{\cV}(g^* \circ P_c, \tig^* \circ \hP_c).
\end{aligned}
\end{equation}
Hence, 
\begin{equation}
\E_{P_1\neq P_2} \left[\textnormal{disc}_{\cV}(g^* \circ P, \tig^* \circ \hP)\right] ~\leq~ \E_{P_c \sim \cD_1} \left[\textnormal{disc}_{\cV}(g^* \circ P_c, \tig^* \circ \hP_c)\right].
\end{equation}
Finally, by Proposition~5 in~\cite{galanti2022on} we obtain that 
\begin{equation}
\begin{aligned}
L_{\hP}(e_{\hP} \circ f \circ \tig, \hiy) ~\leq~ \frac{16V_{f \circ \tig}(\hP_1, \hP_2)}{s(f \circ \tig, \hP_1,\hP_2)}.
\end{aligned}
\end{equation}
\end{proof}

% \subsection{License}
% ModelNet 3D Wharehouse\cite{Wu_2015_CVPR} is extracted from 3D Wharehouse \cite{3dwarehouse}. The license of S3DIS \cite{armeni2017joint} is Apache License 2.0 \footnote{https://github.com/alexsax/2D-3D-Semantics/blob/master/LICENSE}. The license of SemanticKITTI dataset \cite{behley2019iccv} is Creative Commons Attribution-NonCommercial-ShareAlike license \footnote{http://semantic-kitti.org\/dataset.html\#licence}.
\end{document}